\documentclass[5p,times]{elsarticle}
\makeatletter
\def\ps@pprintTitle{%
 \let\@oddhead\@empty
 \let\@evenhead\@empty
 \def\@oddfoot{Published in Pattern Recognition, \url{https://doi.org/10.1016/j.patcog.2025.112441}\hfill}%
 \let\@evenfoot\@oddfoot}
\makeatother

\usepackage{amssymb}
\usepackage{amsmath}
\usepackage{amsthm}
\usepackage{graphicx}
\usepackage{mathtools}
\usepackage{subcaption}
\usepackage{caption}
\usepackage[linesnumbered,ruled,vlined]{algorithm2e}
\usepackage{booktabs}
\usepackage{url}
\usepackage{hyperref}

\usepackage{lineno}
\usepackage[table,xcdraw]{xcolor}

\usepackage{xcolor}
\usepackage{hyperref}
\usepackage{xspace}
\usepackage{comment}
\usepackage{multirow}
\usepackage{multicol}
\usepackage{arydshln}
\usepackage{float}
\usepackage{bbm}
\usepackage{tcolorbox}
\usepackage{booktabs}

\usepackage{caption}
\journal{Pattern Recognition}

\newcommand{\vct}[1]{\ensuremath{\boldsymbol{#1}}}

\newcommand{\set}[1]{\ensuremath{\mathcal{#1}}}

\DeclareMathOperator*{\argmin}{arg\,min}
\DeclareMathOperator*{\argmax}{arg\,max}

\newcommand{\myparagraph}[1]{\noindent \textbf{#1}.}

\newcommand{\mylist}[1]{\noindent \textit{#1}}

\makeatletter
\DeclareRobustCommand\onedot{\futurelet\@let@token\@onedot}
\def\@onedot{\ifx\@let@token.\else.\null\fi\xspace}

\newcommand{\ie}{i.e.,\xspace}
\newcommand{\eg}{e.g.,\xspace}
\newcommand{\aka}{a.k.a.\xspace}

\newcommand{\quotes}[1]{``{\textit{#1}}''\xspace}

\newcommand{\x}{\ensuremath{\vct{x}}\xspace}
\newcommand{\X}{\ensuremath{\set{X}}\xspace}
\newcommand{\Y}{\ensuremath{\set{Y}}\xspace}

\newcommand{\ACA}{LCA\xspace}
\newcommand{\AIA}{AIA\xspace}
\newcommand{\AF}{AF\xspace}
\newcommand{\AUC}{AUC\xspace}
\newcommand{\AUPR}{AUPR\xspace}

\newcommand{\Cifarten}{CIFAR10\xspace}
\newcommand{\Cifarhun}{CIFAR100\xspace}
\newcommand{\Timgnet}{TinyImageNet200\xspace}
\newcommand{\CifartenFiveT}{\Cifarten-5T\xspace}
\newcommand{\CifarhunTenT}{\Cifarhun-10T\xspace}
\newcommand{\CifarhunTwenT}{\Cifarhun-20T\xspace}
\newcommand{\TimageNetFiveT}{T-Imgnet200-5T\xspace}
\newcommand{\TimageNetTenT}{T-Imgnet200-10T}

\newcommand{\BUILD}{BUILD\xspace}
\newcommand{\MORE}{MORE\xspace}
\newcommand{\MoreFw}{MORE-FW\xspace}

\newcommand{\learningrate}{\ensuremath{\alpha}\xspace}
\newcommand{\batchsize}{\ensuremath{s}\xspace}
\newcommand{\epochs}{\ensuremath{N}\xspace}
\begin{document}
\setcitestyle{square} 

\begin{frontmatter}

\title{Buffer-free Class-Incremental Learning with Out-of-Distribution Detection}

\author[unica,sapienza]{Srishti Gupta}
\ead{srishti.gupta@uniroma1.it}

\author[unica]{Daniele Angioni}

\author[unica]{Maura Pintor}

\author[unica]{Ambra Demontis}
\ead{ambra.demontis@unica.it}

\author[cispa]{Lea Schönherr}

\author[unica,unige]{Fabio Roli}

\author[unica]{Battista Biggio}

\affiliation[unica]{organization={DIEE, University of Cagliari}, 
            addressline={Via Marengo 2}, 
            city={Cagliari},
            postcode={09100}, 
            country={Italy}}
\affiliation[sapienza]{organization={DIAG, Sapienza University of Rome}, 
            addressline={Via Ariosto 25},
            city={Rome},
            postcode={00185},
            country={Italy}}
\affiliation[unige]{organization={DIBRIS,University of Genova}, 
            addressline={Via Dodecaneso 35}, 
            city={Genova},
            postcode={16146},
            country={Italy}}
\affiliation[cispa]{organization={CISPA Helmholtz Center for Information Security}, 
            addressline={Stuhlsatzenhaus 5}, 
            city={Saarbrücken},
            postcode={66123}, 
            country={Germany}}
            
\begin{abstract}
Class-incremental learning (CIL) poses significant challenges in open-world scenarios, where models must learn new classes over time without forgetting previous ones and handle inputs from unknown classes that a closed-set model would misclassify. In this paper, we present an in-depth analysis of post-hoc OOD detection methods and investigate their potential to eliminate the need for a memory buffer. When post hoc OOD detection is applied at inference time, we discover that it can effectively replace buffer-based strategies.
We examine the performance of these methods in terms of classification accuracy of seen samples and rejection rates of unseen samples. We show that our approach achieves competitive performance compared to recent multi-head and single-head methods that rely on memory buffers and other buffer-free approaches. The results show that the proposed approach outperforms them in a closed-world setting and detects unseen samples while being significantly resource-efficient. Experimental results on CIFAR-10, CIFAR-100, and Tiny ImageNet support our findings and offer new insights into the design of efficient and privacy-preserving CIL systems for open-world settings.

\end{abstract}


\end{frontmatter}


\section{Introduction}
Recent advances in machine learning (ML) have enabled remarkable successes across diverse practical applications, such as computer vision~\cite{Poyser24-pr} and natural language processing~\cite{hu2024prompting}. 
Despite these successes, deploying ML models in real-world environments exposes two critical challenges. The first stems from the fact that conventional ML models are typically developed under a \textit{closed-world} assumption: the model is trained and evaluated on data drawn from the same underlying distribution, \ie independent and identically distributed (i.i.d.), with the set of possible classes fixed and known a priori. In contrast, the \textit{open-world} setting better reflects real-world conditions, where the model may encounter novel, previously unseen classes at inference time. Under the closed-world assumption, such inputs are inevitably misclassified as belonging to one of the known classes, rather than being rejected, thereby producing errors that undermine reliability~\cite{scheirer2013toward, Condessa17-pr}.
The second challenge lies in continuously extending the model’s capabilities to accommodate new classes. This continual updating process becomes increasingly impractical when considering the substantial financial costs, computational demands, and environmental impact of retraining large models from scratch~\cite{strubell-2019-energy}.

The field of \textit{out-of-distribution} (OOD) detection addresses the first challenge by developing methods to identify and reject inputs originating from unknown classes or distributions from in-distribution (IND) classes at inference time~\cite{hendrycks2017baseline,liu2021energyscoreOOD}. In contrast, \textit{incremental learning} (IL) tackles the second challenge by enabling models to acquire new knowledge over time with minimal or no access to prior training data~\cite{wang2024comprehensive}, thereby avoiding unbounded growth in storage and computational demands.
While the two parallel lines of research have evolved to mitigate these problems separately, there is only limited work targeting the issue jointly.

Acquiring new knowledge without the presence of past data often gives rise to \textit{catastrophic forgetting}, a phenomenon in which gradient-based optimizations overwrite parameters critical to earlier tasks, leading to severe degradation in previously acquired knowledge~\cite{Liu24-pr}. 
One prominent IL scenario is \textit{task-incremental learning} (TIL), where each task typically corresponds to a disjoint set of classes~\cite{delange2021continual}.
State-of-the-art TIL approaches frequently employ \textit{multi-head} architectures~\cite{serra2018overcoming}, which comprise a shared feature-extraction backbone and a set of task-specific classifiers (\textit{heads}).
Here, the shared backbone and current head are updated with ad-hoc strategies that allow near-perfect elimination of catastrophic forgetting~\cite{serra2018overcoming}. During inference, the task identity (task-id) is assumed to be known, enabling the model to route each input to its corresponding head for specialized classification.
For instance, consider a model trained sequentially on two tasks, each containing two classes: $t_0 = \{cat, dog\}$, and $t_1 = \{duck, frog \}$. During inference, an input is first processed by the shared backbone, after which the appropriate classifier head is selected based on the provided task-id.
For example, if the sample belongs to $t_1$, it is routed to the second head, where classification is performed exclusively between the classes \quotes{duck} and \quotes{frog}.

While multi-head architectures have proven effective in mitigating catastrophic forgetting, their reliance on task-id at inference time is a strong and often impractical assumption~\cite{vandeven2019scenarioscontinuallearning}.
A more realistic yet inherently challenging scenario, known as \textit{class-incremental learning} (CIL), requires the models to classify input samples across all previously learned classes without task-id~\cite{masana2022class}.
Recent theoretical and empirical studies~\cite{kim2022theoretical} have shown that solving a CIL problem entails addressing two sub-problems simultaneously: (i) \textit{task-id prediction} (TP), \ie identifying the correct task to which the input belongs, and (ii) \textit{within-task prediction} (WP), \ie predicting the correct class from the predicted task.
Given that TIL models already excel in WP performance, recent works addressed the missing step, \ie strong TP, by integrating OOD detection techniques~\cite{kim2022more, kim2023learnability, lin2024TPL}.
In particular, they use a memory buffer composed of samples from past tasks and train each head to detect OOD samples from different tasks, thus using this mechanism to perform task-id prediction.
Moreover, some works have also shown that, in addition to predicting task ID, this model can also detect never-before-seen samples at inference, making it the ultimate choice for reliable and efficient deployment in an open-world setting~\cite{kim2022more}.
However, relying on memory buffers introduces significant limitations. From a regulatory standpoint, storing past-task data can conflict with privacy laws such as the General Data Protection Regulation GDPR~\footnote{General Data Protection Regulation https://gdpr-info.eu}, especially in sensitive applications like healthcare and finance. 
Employing a buffer facilitates data breaches and, as only a few examples for each class are retained by it and employed to train the model, avoiding forgetting, they will have a high influence on the decision of the classifier, which facilitates ML
poisoning~\cite{cina23-cs} and privacy attacks~\cite{rigaki23-cs}. %
Additionally, replay buffers introduce scalability issues, as increasing the number of tasks significantly amplifies memory and computational requirements, making these models difficult to deploy in resource-constrained environments~\cite{zhou2024class}.  These concerns underscore the need for CIL solutions that eliminate or reduce dependency on stored raw data while retaining strong TP and WP performance.

In this work, we argue that a memory buffer is not necessary for a strong task-id prediction.
Instead, we show that it is sufficient to employ appropriate \textit{post-hoc OOD detection} techniques, \ie methods that detect unknown inputs at inference time, without requiring additional training or architectural changes \cite{hendrycks2017baseline, liu2021energyscoreOOD, lee2018simple, sun2021react, sun2022dice, xu2023scale}.
These models are also capable of detecting unseen OOD samples at inference. We refer to the proposed method as \emph{\textbf{BU}ffer-free \textbf{I}ncremental \textbf{L}earning with OOD \textbf{D}etection} or \BUILD throughout the article.
Extensive experiments conducted on CIFAR-10, CIFAR-100, and Tiny ImageNet demonstrate that \BUILD attains classification accuracy and out-of-distribution rejection rates comparable to recent memory-based multi-head methods such as Multi-head OOD Replay (MORE). Furthermore, \BUILD outperforms single-head approaches, including Dark Experience Replay (DER++) and Prototype Augmentation and Self-Supervision (PASS), the latter exemplar-free method.

To summarize, the main contributions of this work are the following:
\begin{itemize}
    \item We propose \BUILD, a novel buffer-free incremental learning framework that leverages post-hoc OOD detection techniques to enable strong task-id prediction and within-task classification without relying on stored raw data.
    \item We conduct an in-depth analysis of the SOTA multi-head buffered method \MORE, demonstrating that buffers are not essential for OOD detection;
    \item We demonstrate threshold-free rejection performance on unseen samples across varying rejection rates, providing a thorough evaluation of \BUILD’s open-world recognition capabilities.
\raggedright
    \item We conduct comprehensive experiments on standard benchmarks: CIFAR-10, CIFAR-100, Tiny ImageNet, showing that the proposed method attains classification accuracy and OOD detection rates comparable to or exceeding those of recent memory-based and exemplar-free methods.
\par
\end{itemize}

Thus, our analysis unveils scalable and privacy-aware incremental learning by offering an efficient alternative to buffer-dependent approaches, suitable for realistic open-world deployments. The overall performance of the proposed method, benchmarked against the strongest baseline results, is summarized in Table \ref {tab:compact results}. In the following sections, we first present a formalization of the incremental learning problem and the primary methods to mitigate forgetting in~\autoref{sec: prelim}, then present our proposed method in~\autoref{sec:build}. \autoref{sec: experiments} describes the experimental protocol and presents a comprehensive evaluation of our approach; ~\autoref{sec: related_work} discusses related work; and ~\autoref{sec: conclusions} concludes the study with a summary of findings, limitations, and potential future directions.

\begin{table*}[htbp] 
\centering 
\setlength{\tabcolsep}{3pt} 
\renewcommand{\arraystretch}{0.7} 
\resizebox{0.99\linewidth}{!}{%
\begin{tabular}{l|ccccc|ccccc|ccccc}
\toprule

& \multicolumn{5}{c|}{\CifartenFiveT}
& \multicolumn{5}{c|}{\CifarhunTenT}
& \multicolumn{5}{c}{\CifarhunTwenT} \\
\cmidrule(lr){2-16}

 &
\multicolumn{3}{c}{Closed-World} & \multicolumn{2}{c|}{Open-World} &
\multicolumn{3}{c}{Closed-World} & \multicolumn{2}{c|}{Open-World} &
\multicolumn{3}{c}{Closed-World} & \multicolumn{2}{c}{Open-World} \\

\cmidrule(lr){2-4} \cmidrule(lr){5-6} \cmidrule(lr){7-9} \cmidrule(lr){10-11} \cmidrule(lr){12-14} \cmidrule(lr){15-16}

\textbf{Method} & \textbf{\ACA}$\uparrow$ & \textbf{\AIA}$\uparrow$ & \textbf{\AF}$\downarrow$ & \textbf{\AUC}$\uparrow$ & \textbf{\AUPR}$\uparrow$ 
& \textbf{\ACA}$\uparrow$ & \textbf{\AIA}$\uparrow$ & \textbf{\AF}$\downarrow$ & \textbf{\AUC}$\uparrow$ & \textbf{\AUPR}$\uparrow$ 
& \textbf{\ACA}$\uparrow$ & \textbf{\AIA}$\uparrow$ & \textbf{\AF}$\downarrow$ & \textbf{\AUC}$\uparrow$ & \textbf{\AUPR}$\uparrow$  \\

\midrule
\textbf{BUILD (ours)} & 89.86 & 93.7 & 3.65 & 92.16 & 96.31 & 73.47 & 82.16 & 8.61 & 81.28 & 94.57 & 68.98 & 79.11 & 10.37 & 78.56 & 96.48 \\
\midrule
\textbf{PASS} & 83.07 & 89.22 & 12.55 & 82.54 & 92.32 & 70.65 & 79.46 & 14.7 & 79.82 & 93.56 & 69.17 & 78.66 & 16.92 & 78.71 & 95.96 \\
\textbf{$\Delta$PASS} & -6.79 & -4.48 & -8.9 & -9.62 & -3.99 & -2.82 & -2.7 & -6.09 & -1.46 & -1.01 & 0.19 & -0.45 & -6.55 & 0.15 & -0.52 \\
\midrule
\textbf{DER++} & 84.04 & 89.28 & 18.66 & 89.07 & 93.58 & 66.47 & 80.3 & 28.9 & 80.41 & 93.7 & 69.27 & 82.16 & 25.27 & 80.96 & 96.49 \\
\textbf{$\Delta$DER++} & -5.82 & -4.42 & -15.01 & -3.09 & -2.73 & -7.0 & -1.86 & -20.29 & -0.87 & -0.87 & 0.29 & 3.05 & -14.9 & 2.4 & 0.01 \\
\midrule
\textbf{MORE-FW} & 90.24 & 94.06 & 2.04 & 90.9 & 95.95 & 71.73 & 80.47 & 3.13 & 80.86 & 94.43 & 70.26 & 80.03 & 4.43 & 79.18 & 96.26 \\
\textbf{$\Delta$MORE-FW} & 0.38 & 0.36 & 1.61 & -1.26 & -0.36 & -1.74 & -1.69 & 5.48 & -0.42 & -0.14 & 1.28 & 0.92 & 5.94 & 0.62 & -0.22 \\
\midrule
\textbf{MORE} & 90.16 & 93.92 & 7.27 & 89.89 & 95.35 & 70.4 & 81.25 & 22.37 & 81.15 & 94.57 & 69.94 & 81.18 & 22.29 & 80.03 & 96.68 \\
\textbf{$\Delta$MORE} & 0.3 & 0.22 & -3.62 & -2.27 & -0.96 & -3.07 & -0.91 & -13.76 & -0.13 & 0.0 & 0.96 & 2.07 & -11.92 & 1.47 & 0.2 \\
\bottomrule
\end{tabular}%
}
\caption{Performance comparison between the proposed \BUILD framework and the best-performing configurations of baseline methods under both closed-world and open-world scenarios. $\Delta$ denotes the absolute performance difference between \BUILD\ and each baseline, where positive/negative values indicate relative improvement or decline, respectively.}
\label{tab:compact results} 
\end{table*}

\section{Background}
\label{sec: prelim}

In this section, we introduce the main problem formulation of incremental learning in an open-world setting (\autoref{sec:formulation}). To complete the overview of current techniques, we present the main competitor of our method: MORE, which uses a combination of multi-head models and memory buffers to implement a CIL scenario, discuss its limitations (\autoref{sec:more})followed by providing background on post-hoc OOD detection (\autoref{sec: post-hoc det}).

\begin{figure*}[t!]
    
    \centering
    \begin{minipage}{\linewidth}
    \includegraphics[width=\linewidth]{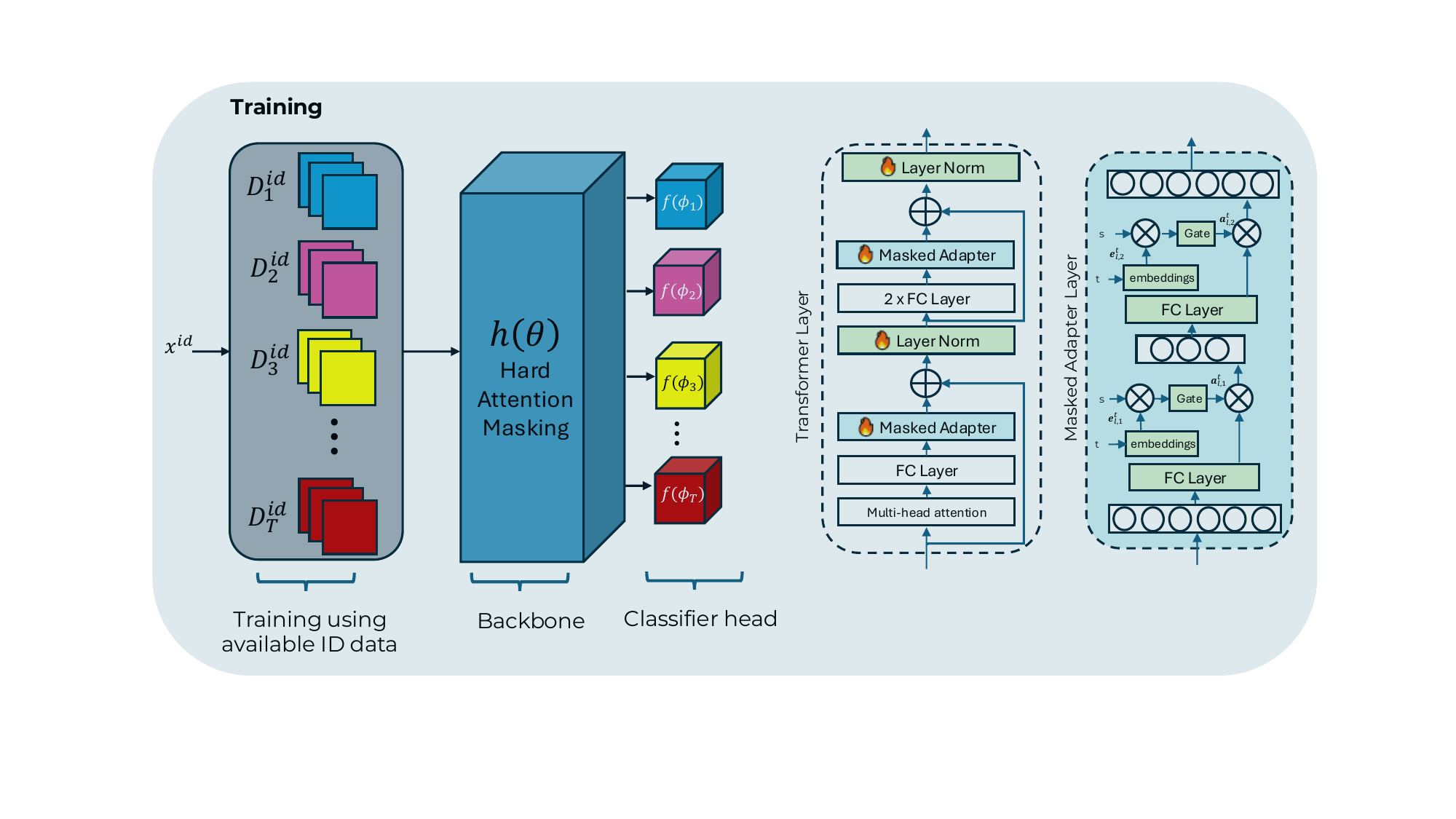}\\[-3.5em]
    \includegraphics[width=\linewidth]{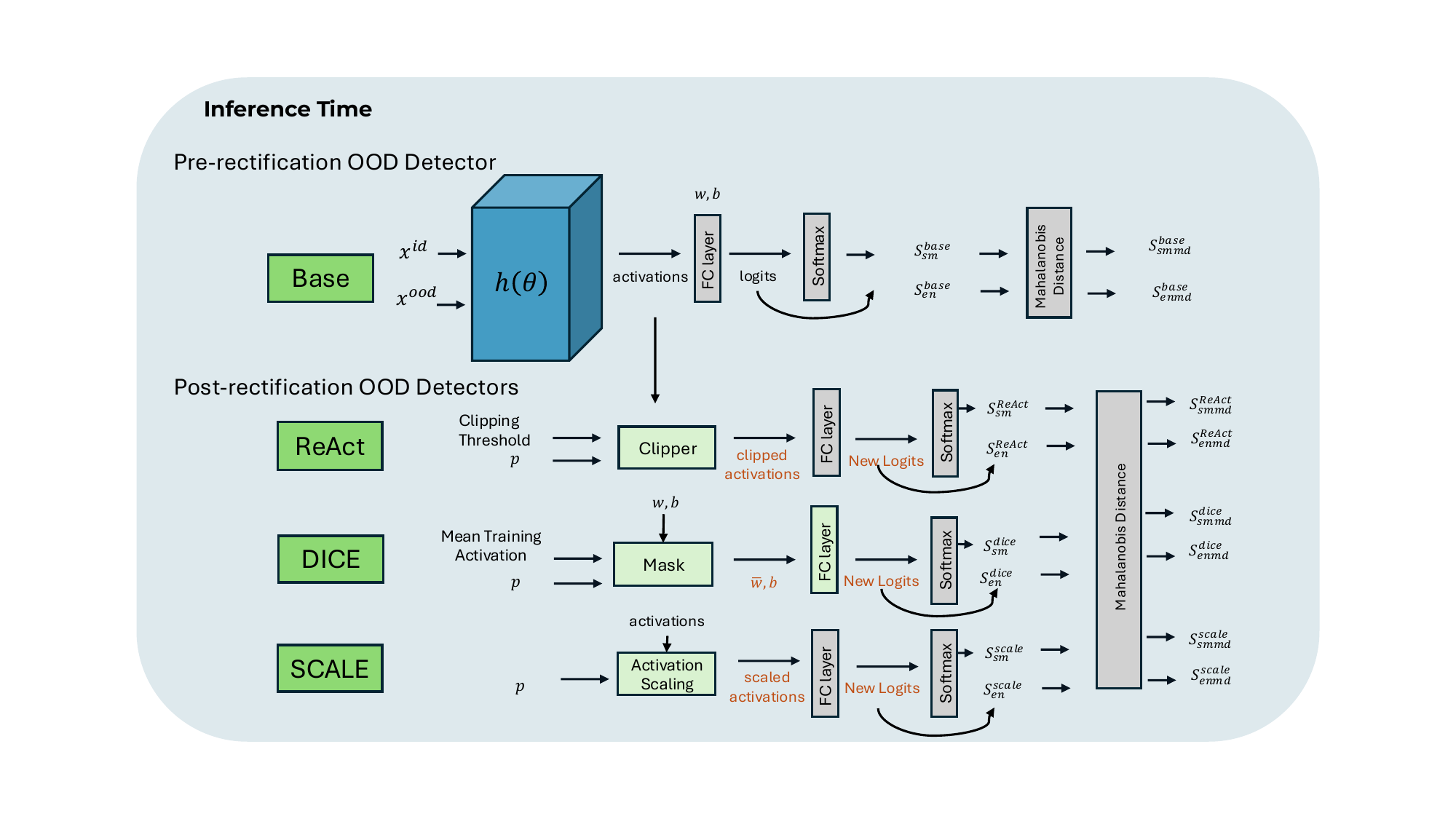}

    \end{minipage}%
    \caption{Graphical representation of \BUILD: a) Training a buffer-free multi-head system with a dedicated head for individual tasks, using masked adapters in the transformer layer, b) Inference time OOD detection and classification using post-hoc detectors, like ReAct, DICE, and SCALE, to increase the separability of IND from OOD samples. The depicted operation is shown per head.}
    \label{fig:graphical}
\end{figure*}

\subsection{Problem Formulation}
\label{sec:formulation}

Let us consider a CIL scenario, where the set of classes $y \in \Y$ must be learned incrementally, divided into a sequence of tasks $t = 1, 2, \ldots, T$.
Each task is characterized by its input space $\X^t$, label space $\Y^t \subseteq \Y$, and dataset $D^t=\{(\x^t_i, y^t_i)\}_{i=1}^{n_t}$ with $n_t$ training instances, where $\x_i^t \in \X^t$ and $y^t_i \in \Y^t$, such that there is no class overlapping between different tasks (\ie $\Y^t \cap \Y^{t'} = \emptyset$ for $t\neq t'$).
We denote $\Y^{\leq t} = \cup_{k=1}^{t} \Y^k$ as the cumulative label space composed of all classes from task $0$ to task $t$. Accordingly, a universal label space $\Y$ can be defined as $\Y^{\leq T}$ where $T$ is the oracle class, \ie, the set of all possible classes.
When learning a new task $t$, the CIL model $f^t$ is required to learn the optimal function $f^t: \X^{\leq t} \rightarrow \Y^{\leq t}$ that maps an input sample to its correct label from either the current task $t$ or the previous.


\subsection{Challenges in Open-World Incremental Learning}
Kim et al.~\cite{kim2022theoretical} demonstrated that, given an input sample $\x$, the output probability of a CIL model can be decomposed into two probabilities, known as (i) \textit{within-task prediction} (WP) and (ii) \textit{task-id prediction} (TP):
\begin{equation}
    P(y^t_j|\x) = \underbrace{P(y^t_j|\x, t)}_\text{WP} \cdot \underbrace{P(t|\x)}_\text{TP} \;
    \label{cil_probs}
\end{equation}

where TP is the probability that $\x$ belongs to one of the tasks the model was incrementally trained upon
and WP is the probability that the correct class label is $j$ given the task-id $t$.\footnote{In the TIL setting, the first term is replaced by knowing exactly the task-id information, which is passed together with the input sample during inference.} 
While WP is a traditional classification task and well-established formulation, TP is a particularly challenging task due to the following reasons:
\begin{itemize}
    \item \emph{Forgetting past knowledge:} Incremental learning methods face the persistent issue of catastrophic forgetting, wherein the parameters of the model adapt to new tasks at the expense of previously learned representations~\cite{rebuffi2017icarl}. This phenomenon is particularly pronounced in training processes that lack explicit mechanisms for knowledge retention, as these methods tend to update parameters globally without consideration for preserving previously-learned tasks~\cite{rebuffi2017icarl}. The most popular IL methods rely on \textit{replay} (or \textit{rehearsal}) strategies. The core idea is to store samples from old tasks in a memory buffer that can be replayed alongside the current task when learning it.
    
    \item \emph{Inter-task Class Separation (ICS):} ICS arises due to the absence of old task data during subsequent trainings, leading to difficulties in defining accurate decision boundaries between classes from different tasks. To alleviate ICS, existing methods employ buffers to store and replay samples from previous classes, facilitating the discrimination between old and new categories. 
 
    \item \emph{Open-world conditions:} A critical challenge in real-world deployment of incremental learning models is the necessity to operate under open-world conditions, where inputs may come from both seen (IND) and unseen (OOD) classes. Standard models tend to erroneously assign these novel inputs to known classes without explicit novelty detection capabilities, resulting in fundamentally flawed predictions and undermining reliability.
\end{itemize}

It is essential to clarify the distinction between these two forms of OOD detection: detecting OOD samples with respect to individual tasks is instrumental for accurate task-id prediction (TP), whereas detecting OOD relative to the entire system corresponds to identifying novel \textit{wild} inputs not encountered during training, which are aimed to be rejected. Although multi-head architectures have demonstrated strong resilience against catastrophic forgetting by isolating network parameters for each task, this work addresses the latter challenges of inter-task class separation and open-world recognition without reliance on memory buffers.

\subsection{Comparison with Recent Works}
\label{sec:more}

A recent approach called Multi-head model for continual learning via OOD REplay (MORE)~\cite{kim2022more} tried to solve the gaps above. MORE combines (i) a multi-head transformer architecture to learn tasks incrementally, (ii) masked adapter modules in the transformer layers to isolate the learning and mitigate forgetting, and (iii) uses replay buffers to perform task-id prediction. 
This is an essential step towards non-forgetting CIL models in open-world conditions. First, using a multi-head architecture creates separate classifiers (heads) for each task, mitigating forgetting by letting the previous heads remain mostly unchanged.
To achieve this, MORE uses OOD detection to enable distinction between tasks without explicit boundaries (\ie, to adapt the multi-head architecture to the CIL setting). They used Mahalanobis Distance as a coefficient to softmax scores as OOD scores. Additionally, MORE combines OOD detection with a buffer to store exemplar samples from previous tasks.
Specifically, it uses the saved samples to provide the classification heads with OOD capabilities for task-id prediction in a CIL setting.

\myparagraph{Multi-head Structure}
Architecturally, a general multi-head setup for incremental learning is composed of (i) a fixed pre-trained backbone $h$ with parameters $\vct \theta$ as a feature extractor, (ii) a set of task-specific trainable adapters with parameters $\vct{\phi}$ inserted in each transformer layer~\cite{houlsby2019parameterefficient}, and (iii) a set of classification heads $f$ with parameters $\vct{w}$.

We now define the backbone that extracts the features using the adapter $\vct{\phi}^t$ specialized on task $t$ as $h$. 
In the following, if not specified, we refer to as $f^t(\x)$ the output of the overall model $f^t(h(\x, \vct{\phi^t, \vct{w}^t}))$ when task $t$ is selected.

Training the model on the task $t$ amounts to finding the optimal configuration of $\vct{\phi}^t$ and $\vct{w}^t$ that minimizes the classification loss, \eg the cross-entropy.

In MORE, the multi-head architecture is trained with a 2-step process.
In the first step, $f^t$ is trained on the dataset $D^t$ and $M$, where the latter is a memory buffer containing samples from previous tasks.
The memory buffer is the leading component responsible for achieving TP, and is used to train an additional logit on $f^t$ to recognize as OOD samples from different tasks. The objective can then be defined as follows:
\begin{equation}
\label{eq:more_fw}
    \argmin_{f^t} \space \sum_{(\x, y) \sim D^t} \set{L}(f^t(\x), y) + \sum_{(\x, y) \sim M} \set{L}(f^t(\x), OOD)
\end{equation}
In a second step, the model performs \textit{back-update}, \ie all previous heads are updated to recognize respective OOD samples from $M$. Back-update takes place after training head $k$ on all preceding heads $j \in \{1, \ldots, k-1\}$. In this case, samples from the future tasks are considered OOD, for \eg when back-updating head $f^2$, samples from tasks $\{t_3, \ldots, T\}$ are considered OOD and are retrieved from the buffer $M$. 
This results in the following optimization process:
\begin{equation}
    \argmin_{w_j} \space \sum_{(\x, y) \sim \tilde{D}^j} \set{L}(f^j(\x), y) + \sum_{(\x, y) \sim \tilde{M}} \set{L}(f^j(\x), y_{OOD})
\end{equation}

where $\tilde{D}^j$ contains the data sampled from task $j$, and $\tilde{M}$ contains randomly selected OOD samples from $M$ after removing the IND samples from task $j$.

We note that in this step, only the parameters of the previous heads are fine-tuned.
Please note we refer to the model derived after the first step as \MoreFw and the models trained after both steps as \MORE and consider both for analysis in \autoref{sec: experiments}.

\myparagraph{Hard Attention Task (HAT) Masking}
HAT isolates subsets of the model parameters for each task to prevent interference among parameters. 
For a layer $l$, and a task $t+1$, the embedding $e_l^t$ will protect the neurons essential for the previous task and should output a value equal to 0 for the neurons necessary for the task $t$ and 1 for the neurons not crucial for that task. Since the step function is not differentiable, a sigmoid function can approximate it, obtaining the attention mask $\vct{a}_l^t=\sigma(se^t_l)$. 
This mask will then be multiplied by the output of layer $l$ during training so that the neurons for the task $t$ are not updated (see masked adapter layer in~\autoref{fig:graphical}(a)). 
This mechanism preserves the attention values for units important for previous tasks while leaving the other neurons to condition on future tasks (see masked adapter block in \autoref{fig:graphical}).

\myparagraph{OOD Replay Mechanism}
At test time, the model can be subjected to either IND or OOD classes: $\x\in\{\x_{ind}, \x_{ood}\}$ such that $\x_{ood}\in\set{D}^k_{ood}$ where $k\neq t$, as expected in the wild. Mahalanobis distance is used as an OOD detector to compute TP. \MORE uses the replay mechanism to expose the trained task heads with samples from other tasks as OOD, akin to outlier exposure.

\myparagraph{Limitations of MORE}
The main limitations of \MORE are related to its usage of a memory buffer. It uses the memory buffer to expose outliers to individual heads, which is less attractive in real-world settings. While using memory buffers has shown a significant boost in performance, they have considerable drawbacks: a) Limited scalability: as the number of classes grows, these methods require additional computation and storage of raw input samples. Even if the memory size is fixed, the ability of the sample sets to represent the original distribution deteriorates; b) Privacy issues: storing data is not always a compliant option, especially for finance and healthcare data. Storing data leaves larger surface area for attacks \cite{biggio18-pr} against ML systems; c) Task-recency bias: typically a dominant task bias is observed towards the more recent classes, which may be caused due to imbalance in current and past samples; d) Memory storage: storing raw samples to buffer consumes enormous memory costs making the process computationally expensive; and e) Compute time: sampling from training data to store in the buffer and updating it for each forward and backward training step costs more computing time in training.

\subsection{Post-hoc OOD Detectors}\label{sec: post-hoc det}

Given a well-trained model $f(\vct{x}, y; \theta)$, post-hoc OOD detectors modify the intermediate results to enhance the separability of IND and OOD samples. They have a low computational cost and require minimal modifications in a plug-and-play manner to the models. 
In this work, we consider the following state-of-the-art post-hoc OOD detection methods in a multi-head setup: 

a) \textit{ReAct}~\cite{sun2021react}: It rectifies the feature vector $h(\vct{x})$ from the penultimate layer of neural networks using the threshold $a$, such that updated output is: $f^{ReAct}(\vct{x}, \theta) = \vct{W}^\top\Bar{h}(\vct{x})+\vct{b}$ such that $\Bar{h}(\vct{x})= \min(h(\vct{x}); a)$. 

b) \textit{DICE}~\cite{sun2022dice}: It performs a directed sparsification based on a measure of contribution and preserves the most important weights in the weight matrix via masking. A masking matrix $\mathcal{M}$ is used to compute the updated function output as: $f^{DICE}(\vct{x}, \theta)=(\mathcal{M} \odot \vct{W})^\top h(\vct{x}) + \vct{b}$.

c) \textit{SCALE}~\cite{xu2023scale}: It scales all the activations equally by a factor $r$ and then thresholds them to identify OOD samples. The updated function output can be computed as: $f^{SCALE}(\vct{x}, \theta)=\vct{W}\cdot (h(\vct{x})\odot \tau) +\vct{b}$ where $\tau = \exp (r)$. The scaling factor $r$ is computed as the ratio of the sum of all the activations versus the sum of the activations with a value higher than the $p^{th}$ percentile of the activations. 

In the considered post-hoc detectors, note that ReAct and SCALE operate on the activation layers, whereas DICE operates on the weight matrices of the classifiers.

\subsection{Scoring Functions}
\label{sec: scoring_func}
We also consider other OOD detectors, using them as scoring functions $\mathcal{S}$, as in~\cite{sun2021react,sun2022dice,xu2023scale}. In particular, we consider the following scoring functions: 

a) \textit{Max Softmax Probability (MSP)}~\cite{hendrycks2017baseline}, which thresholds the softmax outputs to predict the sample as IND or OOD; 

b) \textit{Mahalanobis Distance (MD)}~\cite{lee2018simple}, which assumes that the pre-trained features can be fitted well by a class-conditional Gaussian distribution. 
They define $C$ class-conditional Gaussian distribution with a tied covariance $\Sigma$: $P(f (\vct{x} ) | y=c) = N (f  (\vct{x}) |\mu_c,\Sigma )$, where $\mu_c$ is the mean of the multivariate Gaussian distribution of class $c \in {1,..,C}$ and the class mean (covariance) are estimated on the training samples obtaining $\hat{\mu}_c$ ($\hat{\Sigma}$). 
Then, they define the confidence score $M(\vct{x})$ as the Mahalanobis distance between the test sample $\vct{x}$ and the closest class-conditional distribution: $M(\vct{x})=\max_c -(f (\vct{x})-\hat{\mu}_c)^{T} \hat{\Sigma}^{-1} (f (\vct{x})-\hat{\mu}_c)$.

c) \textit{Energy (EN) Score}~\cite{liu2021energyscoreOOD}, which applies an energy function to a discriminative model, with energy defined as $E(\vct{x};f) = - v \, \log \sum_i^K \epsilon^{ f_i(\vct{x}/v)} $, being $v$ the temperature parameter. High energy can be interpreted as a low likelihood of occurrence. 

\section{Methodology}
\label{sec:build}
We now present our novel framework, named BUILD.
By removing the buffer entirely, we use post-hoc detectors to perform task-id prediction, without changing the TIL training process, which has shown great results.

\subsection{Pipeline}

A high-level overview of \BUILD can be seen in \autoref{fig:graphical}, where the upper box refers to the training time and the lower box to the inference time functionality. \BUILD inherits the same architecture as \MORE 
models as described in \autoref{sec:more}.

\myparagraph{Training} To train the task classifier $t$, the training set $\set{D}^t_{ind}$ with $\mathit{c}$ in-distribution classes: $\forall(\x, y)\in \set{D}^t_{ind}$ and $\{y_1, y_2,..y_c\}\in\set{Y}^t_{ind}$ is used to update the parameters $\vct{\phi}^t$ and $\vct{w}^t$ while leaving the original backbone parameters $\vct \theta$ and the parameters of previously trained task classifiers $\vct{\phi}^{\leq t}$, $\vct{w}^{\leq t}$ unchanged (thanks to masked adapters). 
We also note that \BUILD does not expose outliers during training as done in previous work~\cite{kim2022more, kim2023learnability}, nor does it use time-consuming solutions such as back-updating classification heads.  
In practice, each classifier head is exclusively trained on the corresponding task-specific dataset (see \autoref{alg: build_training}), solely optimizing the left-hand term in~\autoref{eq:more_fw}.

\begin{equation}
\label{eq:build_tr}
    \argmin_{f_{\phi_t}} \space \sum_{(\x, y) \sim D^t_{ind}} \set{L}(f_{\phi_t}(\x), y)
\end{equation}

\myparagraph{Inference} At test time, after training task $t$, the model can be subjected to either IND or OOD classes: $\x\in\{\x_{id}, \x_{ood}\}$ such that $\x_{ood}\in\set{D}^{t'}_{ood}$ where $t'\neq t$, as expected in the wild. Task-id prediction is performed when the system operates in a closed-world setting: $\x\in\set{D}^{\leq t}$. 
This is done by maximizing the OOD scores derived from each head to assign the task-id, consequently the class within the predicted task (see~\autoref{alg: build_inference}). 
However, when the system is deployed in an open-world setting, OOD scores are evaluated based on their probability of being rejected as OOD.

\begin{algorithm}
\caption{Training \BUILD}
\label{alg: build_training}
\SetKwInOut{Input}{Input}
\SetKwInOut{Output}{Output}

\Input{sequence of training data tasks $\set{D}_{train} = \cup_{t=1}^T\set{D}^t_{train}$, learning rate $\lambda$, learnable parameters: backbone model $\theta$, task embeddings $\vct{e}$, task heads $\phi$}

\For{each task data $\set{D}^t$} {
    // train model\;
    \For{a batch $(\vct{x}^t_i, y)$ in $\set{D}^t$}{
    Compute $\set{L}$ from~\autoref{eq:build_tr} and model gradients\;
    Update model parameters $\theta$ using HAT~\cite{serra2018overcoming} \;
    Update embeddings $\vct{e}^t \gets \vct{e}^t - \lambda \nabla_{\vct{e}^t} \set{L}$\;
    Update task heads $\phi^t\gets \phi^t-\lambda \nabla_{\phi^t} \set{L}$\;
    }
    \If {last epoch}{
    // save metrics \;
    Compute class and task centroids: $\mu_c$, $\mu_{t}$, $\sigma_c$ $\sigma_{t}$\;
    }
    }
\end{algorithm}

\begin{algorithm}
\caption{Inferencing \BUILD}
\label{alg: build_inference}
\SetKwInOut{Input}{Input}
\SetKwInOut{Output}{Output}

\Input{test samples from $\set{D}_{test}$, scorer $\mathcal{S}$, detectors $\Gamma$}
\For {each scorer-detector combination $\mathcal{S}-\Gamma$}{
    \For {each task $t$}{
        Compute activations $\vct{z}_t$ using ~\autoref{eq: activation}\;
        Compute post-hoc rectifications $f_t^{\Gamma}$ using~\autoref{eq: rectified_features}~\autoref{eq: rectified_features_dice}\;
        }
    // concatenate outputs for closed-world performance\; 
    Predict task-id $\hat{t}$ using~\autoref{eq: task-id pred}\;
    Predict class-id $\hat{y}_{\hat{t}}$ using ~\autoref{eq: class-id pred}\; 
}
Compute evaluation metrics\;

\end{algorithm}

\subsection{Combining OOD Detectors and Scoring Functions for OOD+CIL}
The OOD scores are the model outputs used to evaluate the OOD-ness of the sample. Based on the OOD detector, the OOD scores can be derived at any output stage: activations, logits, or probability space. Following our framework, we can choose $\Gamma$ as a post-hoc OOD detector, such as ReAct, DICE, or SCALE. In this case, $\Gamma$ wraps the last part of the network, starting from the penultimate layer of the task classifier $t$ to extract activations:
\begin{equation}
\label{eq: activation}
    \vct{z}_t = h_t(\x)
\end{equation}

As shown in the inference block in the \autoref{fig:graphical}. 

The OOD detector $\Gamma$ is then used to obtain the new activations $\Bar{\vct{z}_t} = \Gamma(\vct{z}_t)$.
The modified features are then used to perform OOD detection:
\begin{equation}
    f_t^{\Gamma}(\x, \theta) = \vct{w}_t \times \Bar{\vct{z}_t} + b_t
    \label{eq: rectified_features}
\end{equation}
Since DICE performs modification in the weights of the last fully-connected layer instead of the output space, we can rewrite \autoref{eq: rectified_features} for DICE as:  
\begin{equation}
    f_t^{\Gamma}(\x, \theta) = M*\vct{w}_t \times \vct{z}_t + b_t \,
    \label{eq: rectified_features_dice}
\end{equation}
where $M$ are the masks for important units from the classification layer. 

Depending on the scoring function $\mathcal{S}$, the final scores can be represented as $\mathcal{S}(f_t^{\Gamma}(\x, \theta))$. In this work, we analyze the performance of the OOD detectors across four scorers \aka $\mathcal{S}$ (mentioned in \autoref{sec: scoring_func}): (i) Softmax (SM) scores: Applying softmax at the final layer and getting the maximum for final output, basically MSP; (ii) MD as coefficient to SM (SMMD) scores (as in~\cite{kim2022more}); (iii) Energy (EN) scores, mapping the logit outputs from the network $f(\x)$ to a scalar value; and (iv) MD applied to EN (ENMD) scores, using the same mean and covariance parameters from the training data. 

After we get these scores, we can evaluate the model for closed-world and open-world settings. For a given detector $\Gamma$, we can also predict task-id :
\begin{equation}
\label{eq: task-id pred}
    \hat{t} = \argmax \bigoplus_{1\leq t \leq T} \mathcal{S}(f_t^{\Gamma}(\x, \theta))
\end{equation}
where $\bigoplus$ is the concatenation over the output space. Further, class-id can be predicted from the predicted task-id using the original activations $\vct{z}_t$, which is guaranteed to give identical classification accuracy~\cite{sun2021react}

\begin{equation}
\label{eq: class-id pred}
    \hat{y}_{\hat{t}} = \argmax\bigoplus_{1\leq c \leq C} \sigma(f_{\hat{t}}(\vct{x}, \theta)
\end{equation}
where $f_{\hat{t}}(\vct{x}, \theta)=\vct{w}_{\hat{t}} \times \vct{z}_{\hat{t}} + b_{\hat{t}}$ and $\sigma$ is the softmax operator.

\section{Experiments}
\label{sec: experiments}

In this section, we first define the experimental setup, evaluation metrics considered for closed-world and open-world scenarios, and make comparisons with \MORE as the baseline, followed by other low-resource methods like DER++ and PASS (exemplar-free). Furthermore, we provide a detailed analysis of the results across all configurations of scoring functions and OOD detectors, alongside assessing computational resource usage and training time requirements.

\subsection{Experimental Setup}

\myparagraph{Baseline} We consider MORE as our baseline. In our experiments, we leveraged the implementation provided by its authors. We analyze the version proposed by the authors and the intermediate variant \MoreFw, and then the method \BUILD under different OOD methods, including pre- and post-rectification and scoring functions. We finally do an ablation study, where we consider the best-performing post-hoc OOD detector for each technique.

\myparagraph{Backbone} We use the following architecture: DeiT-S/16~\cite{pmlr-v139-touvron21a} with 2-layer adapter module~\cite{houlsby2019parameterefficient} with 64-dimensional latent space in each transformer layer. We borrowed the checkpoints of the backbone transformer from the baseline work, where Vision Transformer(ViT) is pre-trained on ImageNet classes after removing 389 classes that are similar to the classes in CIFAR datasets to avoid data leaking at train time~\cite{kim2022more}. We plug OOD detectors into the output space.

\myparagraph{Evaluation Metrics} We group the evaluation metrics into closed- and open-world performance metrics.

\mylist{Closed-world Performance}. The metrics here evaluate the model's performance on seen classes and how well the model can predict the correct class in the incremental learning setup.
\begin{itemize}
    \item Last Classification Accuracy (\ACA). Accuracy of all seen classes after the last task $T$ is trained: ${LCA} = A^T$.
    \item Average Incremental Accuracy (AIA). Average of all the accuracies computed after each incremental step: $AIA =1/T \sum_{t=1}^T A^t$.
    \item Average Forgetting (AF): Forgetting refers to the performance decline of the task classifier after training task $t$ compared to the performance when the task head was first learned. The reported results are averaged considering all seen tasks.
    $\mathcal{F}=1/(T-1) \sum_{t=1}^{T-1}\{A^{init}_t - A_t^T\}$.
\end{itemize}
 
\mylist{Open-world Performance}. The metrics used here are used to assess the performance of the models on unseen classes and how distinguishable the scores of seen vs unseen samples are for different detectors. 
We also note that comparing detectors is not as straightforward as evaluating accuracy. 
We consider two classes for detection: positive (IND) and negative (OOD). 
An important consideration is the shift in the IND to OOD data ratio after each incremental training. 
For \eg in \CifartenFiveT, after the first training, IND is 20\% of the test data, while OOD is 80\%, which changes by the time we have trained the second last task to: IND to 80\% and OOD to 20\%. Therefore, we average the metrics after each training until the second-to-last model.
\begin{itemize}
    \item Average Area Under the Curve (\AUC). AUC \cite{Bradley97-pr} is a threshold-independent metric that computes the area under the ROC curve, i.e., how TPR varies against an increasing FPR.
    During incremental training, the ratio of the two classes continuously changes; thus, we calculate the \AUC after each incremental step and average it.
    \item Average Area Under Precision-Recall (\AUPR). As mentioned in ~\cite{saito2015_aupr}, AUPR is a suitable metric for imbalanced datasets where positive and negative classes can have differing base rates. We average the \AUPR as we do for the \AUC.
\end{itemize}

\myparagraph{Datasets} We run our experiments on three image classification datasets: \Cifarten, \Cifarhun, and \Timgnet.
\Cifarten is composed of $60,000$ $32\times32$ RGB images of $10$ classes. Per class, there are $5000$ samples for training and $1000$ for testing. We consider \CifartenFiveT where a sequence of $5$ tasks, composed of $2$ classes each, is trained, without changing the original class order, \eg, task $t_0$ comprises samples from classes 0, 1; task $t_1$ samples from class 2, 3, etc.
\Cifarhun is composed of $60,000$ $32\times32$ RGB images of 100 classes with $500$ samples for training and $100$ samples for testing in each class. We conduct the experiments employing two different training settings: (i) \CifarhunTenT: sequence of $10$ tasks: $t_0, .., t_9$, each task composed of $10$ classes, is used to train the model; and (ii) \CifarhunTwenT: sequence of $20$ tasks: $t_0, .., t_{19}$, each task composed of $5$ classes is used to train the model.
\Timgnet is composed of $120,000$ $64\times64$ color images of 200 classes with $500$ samples for training, $50$ for validation and $50$ for testing in each class. Following previous works~\cite{kim2022more}, we use a validation dataset for testing since the test set doesn't have labels.

\myparagraph{Training Details} We train the task-specific adapters, normalization layers, and classification heads using the stochastic gradient descent (SGD) optimizer for \epochs epochs with learning rate \learningrate, and batch size \batchsize. 

\noindent \emph{\CifartenFiveT:} We set \epochs= 20, \learningrate= 0.005 and \batchsize= 64 for both settings. The bottleneck size of the adapter is 64. Additionally, for \MORE, we set a buffer size of 200 and an additional 10 epochs for the back-update procedure.

\noindent \textit{\CifarhunTwenT:} We set \epochs= 40, \learningrate= 0.005 and \batchsize= 64 for both settings. The bottleneck size of the adapter is 64. For \MORE, we use the default setting with a buffer size of 2000 and an additional 10 epochs for back-update.

\noindent \textit{\CifarhunTenT:} We used the same hyperparameters as \CifarhunTwenT except for \learningrate= 0.001, as we have more classes each training round.

\noindent \textit{\TimageNetFiveT:} We used \epochs=25, \learningrate=0.005, \batchsize=64. The bottleneck size is 128. For \MORE experiments, the buffer size is 2000 samples in the memory. As in the original work, they do not perform backupdate for this dataset; therefore, we only perform forward pass and refer to them as \MoreFw.

\noindent \textit{\TimageNetTenT: } We used similar settings as \TimageNetFiveT except for \epochs=20.

For OOD detection at inference, we save specific parameters at the training time that are used later \eg the mean and standard deviation of each class and task to compute Mahalanobis Distance, mean activations, and clipping threshold employed by DICE and ReAct, respectively. SCALE does not require any parameters for the training time. We use the percentile value $p$ recommended in the original paper for rectification, \ie 90, 85, 85 for ReAct, DICE, and SCALE, respectively.

\myparagraph{Task-id Prediction} At inference time, we first compute task-id prediction using OOD detectors. As shown in \autoref{fig:graphical}, OOD detectors are deployed at different stages of the output space depending on the detector type. We first compute the pre-rectification detection method (Base) from the baseline method, \ie Softmax scores with and without the combination with Mahalanobis Distance (MD), referred to as SM and SMMD. At this \textit{Base} stage, we calculate Energy scores with and without the combination of MD, called EN and ENMD, respectively.
We then used three post-rectification OOD detectors to predict task-id: (i) ReAct, (ii) DICE, and (iii) SCALE using EN and ENMD scores. For a fair comparison, we analyze the results of \BUILD and \MORE for both pre- and post-rectification methods.

\definecolor{lightgray}{gray}{0.9}
\definecolor{lightblue}{RGB}{230, 245, 255}

\begin{table*}[htbp] 
\centering 
\resizebox{0.99\linewidth}{!}{%
\begin{tabular}{lll|ccccc|ccccc|ccccc}
\toprule

\multicolumn{3}{c|}{} 
& \multicolumn{5}{c|}{\CifartenFiveT}
& \multicolumn{5}{c|}{\CifarhunTenT}
& \multicolumn{5}{c}{\CifarhunTwenT} \\
\cmidrule(lr){4-18}

\multicolumn{3}{c|}{} &
\multicolumn{3}{c}{Closed-World} & \multicolumn{2}{c|}{Open-World} &
\multicolumn{3}{c}{Closed-World} & \multicolumn{2}{c|}{Open-World} &
\multicolumn{3}{c}{Closed-World} & \multicolumn{2}{c}{Open-World} \\

\cmidrule(lr){4-6} \cmidrule(lr){7-8} \cmidrule(lr){9-11} \cmidrule(lr){12-13} \cmidrule(lr){14-16} \cmidrule(lr){17-18}

 &  &  
& \textbf{\ACA}$\uparrow$ & \textbf{\AIA}$\uparrow$ & \textbf{\AF}$\downarrow$ & \textbf{\AUC}$\uparrow$ & \textbf{\AUPR}$\uparrow$ 
& \textbf{\ACA}$\uparrow$ & \textbf{\AIA}$\uparrow$ & \textbf{\AF}$\downarrow$ & \textbf{\AUC}$\uparrow$ & \textbf{\AUPR}$\uparrow$ 
& \textbf{\ACA}$\uparrow$ & \textbf{\AIA}$\uparrow$ & \textbf{\AF}$\downarrow$ & \textbf{\AUC}$\uparrow$ & \textbf{\AUPR}$\uparrow$  \\

\midrule
\multirow{16}{*}{\rotatebox{90}{\textbf{\MoreFw}}} 
& \multirow{4}{*}{\rotatebox{90}{\textbf{SM}}}
 & \textbf{Base} & 88.51 & 92.23 & 2.54 & 90.71 & 95.93 & 64.45 & 74.16 & 2.04 & 79.71 & 93.88 & 64.44 & 75.0 & 2.56 & 78.98 & 96.22 \\
 & & \textbf{React} & 88.05 & 91.86 & 2.44 & 90.11 & 95.65 & 65.36 & 75.05 & 2.36 & 79.87 & 93.98 & 64.79 & 75.37 & 2.71 & 79.11 & 96.28 \\
 & & \textbf{Dice} & 86.3 & 91.12 & 3.74 & 89.17 & 95.34 & 52.76 & 66.3 & 1.1 & 76.99 & 92.78 & 59.33 & 71.28 & 2.66 & 76.4 & 95.52 \\
 & & \textbf{Scale} & 88.66 & 92.31 & 2.52 & \textbf{90.9} & \textbf{95.95} & 64.59 & 74.21 & 2.09 & 79.98 & 93.97 & 64.61 & 75.07 & 2.55 & \textbf{79.18} & \textbf{96.26} \\
\cline{4-18}
& \multirow{4}{*}{\rotatebox{90}{\textbf{SMMD}}}
 & \textbf{Base} & 86.06 & 91.16 & 1.79 & 85.16 & 93.01 & \textbf{71.73} & \textbf{80.47} & \textbf{3.13} & 80.65 & 94.36 & 70.15 & 79.98 & 4.54 & 78.15 & 96.32 \\
 & & \textbf{React} & 85.96 & 91.17 & 1.99 & 85.48 & 93.13 & 71.6 & 80.17 & 3.34 & 80.78 & 94.43 & 69.93 & 79.95 & 4.47 & 78.21 & 96.35 \\
 & & \textbf{Dice} & 86.97 & 91.81 & 2.16 & 86.66 & 93.78 & 59.16 & 71.0 & 1.29 & 79.68 & 94.2 & 67.21 & 77.78 & 3.42 & 78.57 & 96.41 \\
 & & \textbf{Scale} & 86.18 & 91.25 & 1.85 & 85.31 & 93.07 & 71.78 & 80.45 & 3.09 & \textbf{80.86} & \textbf{94.43} & \textbf{70.26} & \textbf{80.03} & \textbf{4.43} & 78.28 & 96.35 \\
\cline{4-18}
& \multirow{4}{*}{\rotatebox{90}{\textbf{EN}}}
 & \textbf{Base} & 77.53 & 87.68 & 14.96 & 85.24 & 93.44 & 38.26 & 55.26 & 27.93 & 72.08 & 91.47 & 25.11 & 41.12 & 27.05 & 67.2 & 93.73 \\
 & & \textbf{React} & 77.71 & 86.97 & 14.01 & 83.56 & 92.86 & 40.9 & 55.96 & 25.34 & 72.48 & 91.56 & 27.57 & 43.38 & 26.55 & 68.22 & 94.11 \\
 & & \textbf{Dice} & 79.99 & 89.49 & 7.71 & 82.5 & 92.3 & 5.89 & 27.92 & 40.98 & 62.03 & 86.31 & 8.35 & 25.48 & 29.55 & 61.19 & 91.65 \\
 & & \textbf{Scale} & 76.89 & 87.52 & 16.16 & 85.86 & 93.55 & 37.47 & 54.84 & 29.03 & 72.29 & 91.51 & 24.47 & 40.51 & 27.55 & 67.23 & 93.68 \\
\cline{4-18}
& \multirow{4}{*}{\rotatebox{90}{\textbf{ENMD}}}
 & \textbf{Base} & 90.15 & 94.02 & 2.06 & 90.16 & 95.85 & 69.05 & 78.25 & 10.3 & 80.36 & 94.37 & 65.91 & 75.8 & 10.08 & 78.78 & 96.56 \\
 & & \textbf{React} & 89.71 & 93.71 & 2.24 & 89.16 & 95.54 & 69.1 & 77.84 & 9.76 & 80.03 & 94.28 & 65.27 & 75.6 & 10.46 & 78.62 & 96.59 \\
 & & \textbf{Dice} & 88.62 & 93.35 & 2.04 & 88.46 & 95.11 & 54.37 & 67.18 & 23.17 & 74.15 & 92.0 & 56.94 & 69.27 & 16.2 & 75.18 & 95.76 \\
 & & \textbf{Scale} & \textbf{90.24} & \textbf{94.06} & \textbf{2.04} & 90.42 & 95.87 & 69.25 & 78.25 & 10.36 & 80.59 & 94.43 & 65.81 & 75.79 & 10.23 & 78.91 & 96.58 \\
\bottomrule
 \rowcolor{lightblue}
 & & \textbf{Avg.} & 85.47 & 91.23 & 5.02 & 87.43 & 94.4 & 56.61 & 68.58 & 12.21 & 77.03 & 93.0 & 54.38 & 66.34 & 11.56 & 75.14 & 95.52 \\
\bottomrule
\bottomrule

\multirow{16}{*}{\rotatebox{90}{\textbf{\MORE}}} 
& \multirow{4}{*}{\rotatebox{90}{\textbf{SM}}}
 & \textbf{Base} & 88.96 & 92.95 & 11.22 & 89.88 & 95.32 & 69.54 & 80.82 & 23.63 & 79.6 & 93.77 & 68.94 & 80.59 & 24.25 & 78.52 & 96.21 \\
 & & \textbf{React} & 88.98 & 93.14 & 11.0 & 89.62 & 95.15 & 69.08 & 80.52 & 23.72 & 79.47 & 93.77 & 68.79 & 80.43 & 24.28 & 78.36 & 96.21 \\
 & & \textbf{Dice} & 85.95 & 91.45 & 14.99 & 88.9 & 94.97 & 55.92 & 73.76 & 39.1 & 72.61 & 91.61 & 61.79 & 76.25 & 32.8 & 74.03 & 95.32 \\
 & & \textbf{Scale} & 88.93 & 92.93 & 11.22 & 89.86 & 95.24 & 69.53 & 80.84 & 23.67 & 79.7 & 93.82 & 68.98 & 80.6 & 24.23 & 78.57 & 96.23 \\
\cline{4-18}
& \multirow{4}{*}{\rotatebox{90}{\textbf{SMMD}}}
 & \textbf{Base} & 90.15 & 93.9 & 7.24 & 86.52 & 93.36 & \textbf{70.4} & \textbf{81.25} & \textbf{22.37} & 81.06 & 94.52 & 69.94 & 81.16 & 22.25 & 79.98 & 96.67 \\
 & & \textbf{React} & 90.15 & 93.91 & 6.98 & 86.53 & 93.39 & 70.07 & 81.01 & 22.42 & 80.94 & 94.53 & 69.73 & 81.04 & 22.31 & 79.82 & 96.66 \\
 & & \textbf{Dice} & 89.06 & 93.35 & 9.38 & 87.66 & 94.03 & 57.85 & 74.85 & 37.03 & 74.7 & 92.76 & 62.77 & 77.1 & 31.55 & 76.16 & 96.13 \\
 & & \textbf{Scale} & \textbf{90.16} & \textbf{93.92} & \textbf{7.27} & 86.67 & 93.43 & 70.36 & 81.25 & 22.4 & \textbf{81.15} & \textbf{94.57} & \textbf{69.94} & \textbf{81.18} & \textbf{22.29} & \textbf{80.03} & \textbf{96.68} \\
\cline{4-18}
& \multirow{4}{*}{\rotatebox{90}{\textbf{EN}}}
 & \textbf{Base} & 42.98 & 64.67 & 60.79 & 80.21 & 91.1 & 4.7 & 30.93 & 25.07 & 65.83 & 88.97 & 1.59 & 18.08 & 25.18 & 61.4 & 92.37 \\
 & & \textbf{React} & 44.68 & 64.65 & 57.76 & 79.17 & 90.7 & 5.75 & 31.72 & 25.26 & 65.87 & 88.99 & 1.96 & 19.25 & 26.27 & 61.79 & 92.64 \\
 & & \textbf{Dice} & 46.95 & 69.79 & 55.55 & 81.14 & 91.46 & 1.98 & 17.93 & 12.02 & 59.71 & 85.23 & 0.42 & 12.07 & 15.46 & 58.18 & 90.89 \\
 & & \textbf{Scale} & 42.44 & 64.36 & 61.66 & 80.84 & 91.25 & 4.53 & 30.88 & 25.26 & 66.16 & 89.12 & 1.48 & 17.9 & 24.98 & 61.59 & 92.37 \\
\cline{4-18}
& \multirow{4}{*}{\rotatebox{90}{\textbf{ENMD}}}
 & \textbf{Base} & 84.72 & 90.45 & 14.23 & 89.83 & 95.43 & 51.35 & 67.36 & 11.13 & 76.64 & 93.4 & 44.45 & 61.02 & 23.29 & 74.51 & 96.0 \\
 & & \textbf{React} & 84.75 & 90.32 & 14.04 & 89.48 & 95.37 & 51.92 & 67.35 & 10.58 & 76.57 & 93.4 & 45.2 & 61.41 & 22.31 & 74.53 & 96.04 \\
 & & \textbf{Dice} & 83.97 & 90.2 & 14.94 & 88.88 & 94.78 & 38.46 & 55.24 & -1.28 & 69.11 & 90.2 & 34.87 & 52.68 & 18.52 & 69.42 & 94.78 \\
 & & \textbf{Scale} & 84.75 & 90.48 & 14.2 & \textbf{89.89} & \textbf{95.35} & 51.26 & 67.43 & 11.43 & 76.89 & 93.48 & 44.08 & 60.77 & 23.57 & 74.63 & 96.01 \\
\bottomrule
 \rowcolor{lightblue}
 & & \textbf{Avg.} & 76.72 & 85.65 & 23.28 & 86.57 & 93.77 & 46.42 & 62.7 & 20.86 & 74.13 & 92.01 & 44.68 & 58.85 & 23.97 & 72.6 & 95.08 \\
\bottomrule
\bottomrule

\multirow{16}{*}{\rotatebox{90}{\textbf{\BUILD}}}
& \multirow{4}{*}{\rotatebox{90}{\textbf{SM}}}
 & \textbf{Base} & 87.2 & 91.57 & 6.36 & 92.09 & 96.36 & 62.9 & 74.17 & 13.36 & 79.29 & 93.54 & 59.41 & 71.96 & 14.23 & 78.21 & 96.17 \\
 & & \textbf{React} & 86.63 & 91.27 & 5.79 & 91.56 & 96.04 & 63.68 & 74.85 & 12.69 & 79.32 & 93.57 & 59.77 & 72.41 & 14.22 & 78.11 & 96.16 \\
 & & \textbf{Dice} & 84.98 & 90.35 & 7.07 & 91.42 & 95.96 & 57.02 & 70.81 & 15.43 & 77.11 & 92.47 & 52.83 & 66.55 & 15.27 & 75.66 & 95.48 \\
 & & \textbf{Scale} & 87.23 & 91.61 & 6.44 & \textbf{92.16} & \textbf{96.31} & 62.96 & 74.3 & 13.53 & 79.6 & 93.65 & 59.29 & 71.95 & 14.59 & 78.39 & 96.22 \\
\cline{4-18}
& \multirow{4}{*}{\rotatebox{90}{\textbf{SMMD}}}
 & \textbf{Base} & 84.64 & 90.01 & 3.04 & 84.49 & 92.65 & 71.48 & 80.44 & 8.99 & 79.51 & 94.04 & 66.86 & 77.28 & 9.42 & 75.06 & 95.76 \\
 & & \textbf{React} & 84.57 & 90.01 & 3.19 & 84.71 & 92.74 & 71.65 & 80.42 & 8.78 & 79.85 & 94.18 & 67.26 & 77.52 & 9.48 & 75.42 & 95.83 \\
 & & \textbf{Dice} & 85.38 & 90.55 & 3.56 & 86.29 & 93.53 & 68.17 & 78.68 & 10.24 & 80.14 & 94.23 & 67.01 & 77.18 & 9.49 & 76.06 & 95.95 \\
 & & \textbf{Scale} & 84.69 & 90.05 & 3.01 & 84.62 & 92.71 & 71.63 & 80.54 & 9.02 & 79.73 & 94.11 & 66.93 & 77.35 & 9.48 & 75.18 & 95.78 \\
\cline{4-18}
& \multirow{4}{*}{\rotatebox{90}{\textbf{EN}}}
 & \textbf{Base} & 88.06 & 92.59 & 6.52 & 87.93 & 94.24 & 69.41 & 79.48 & 10.79 & 79.39 & 93.53 & 62.9 & 75.26 & 13.87 & 75.32 & 95.52 \\
 & & \textbf{React} & 86.93 & 91.93 & 5.84 & 87.28 & 94.06 & 68.65 & 78.86 & 10.66 & 79.38 & 93.55 & 63.12 & 75.45 & 13.56 & 75.89 & 95.65 \\
 & & \textbf{Dice} & 85.63 & 91.23 & 7.87 & 85.61 & 92.95 & 59.07 & 72.47 & 15.13 & 71.96 & 90.19 & 55.46 & 69.61 & 15.03 & 70.4 & 94.1 \\
 & & \textbf{Scale} & 87.89 & 92.47 & 6.55 & 87.98 & 94.07 & 69.29 & 79.4 & 11.28 & 79.6 & 93.58 & 62.88 & 75.32 & 14.37 & 75.46 & 95.54 \\
\cline{4-18}
& \multirow{4}{*}{\rotatebox{90}{\textbf{ENMD}}}
 & \textbf{Base} & 89.81 & 93.66 & 3.65 & 90.35 & 95.63 & \textbf{73.47} & \textbf{82.16} & \textbf{8.61} & 81.02 & 94.48 & \textbf{68.98} & \textbf{79.11} & \textbf{10.37} & 78.35 & 96.43 \\
 & & \textbf{React} & 88.74 & 92.97 & 3.44 & 89.71 & 95.43 & 72.91 & 81.7 & 8.46 & 80.7 & 94.4 & 68.47 & 78.89 & 10.58 & 78.21 & 96.42 \\
 & & \textbf{Dice} & 88.6 & 92.8 & 3.9 & 88.94 & 94.81 & 69.4 & 79.6 & 10.37 & 77.46 & 93.01 & 66.24 & 77.26 & 10.17 & 76.03 & 95.85 \\
 & & \textbf{Scale} & \textbf{89.86} & \textbf{93.7} & \textbf{3.65} & 90.4 & 95.53 & 73.37 & 82.09 & 8.82 & \textbf{81.28} & \textbf{94.57} & \textbf{68.97} & \textbf{79.11} & \textbf{10.61} & \textbf{78.56} & \textbf{96.48} \\
\bottomrule
 \rowcolor{lightblue}
 & & \textbf{Avg.} & 86.93 & 91.67 & 4.99 & 88.47 & 94.56 & 67.82 & 78.12 & 11.01 & 79.08 & 93.57 & 63.52 & 75.14 & 12.17 & 76.27 & 95.83 \\
\bottomrule
\end{tabular}%
}
\caption{Results for model: MORE, \MoreFw, and BUILD on the CIFAR-10 and CIFAR-100 datasets, using different scorers and detectors, along with closed-world and open-world metrics. 
For each model and dataset, the row with the highest AIA (ACA in case of ties) is highlighted in bold for ACA, AIA, and AF, while the row with the highest AUC (AUPR in case of ties) is highlighted in bold for AUC and AUPR.
}
\label{tab:full_results} 
\end{table*}

\subsection{Evaluation Protocol} 
We evaluate the (i) Closed-World performance, namely the classification of seen samples using ACA, AIA, and AF, and (ii) Open-World performance, namely the discriminative power on unseen samples using AUC and AUPR. 

\myparagraph{Closed-World Evaluation} 
    For this evaluation, only test samples from the seen distribution are evaluated. At any given incremental step $k$, for any task $t\in \{1, 2, .., k\}$, all test samples belonging to tasks other than $t$ are considered \textit{near} OOD~\cite{yang2022openood}, i.e., $D_t = \bigcup_{\substack{i = 1 \\ i \ne t}}^{k} \mathcal{D}_i$, and the samples from $t$ as In-Distribution (IND). We evaluate all classifier heads until the last trained head $\hat{f_1}, .., \hat{f}_k$ using the detectors mentioned in \autoref{sec: post-hoc det}. The respective detectors produce an OOD score for each class, which is concatenated and compared to determine the task ID. We are basically choosing the class with the least OOD score or highest ID score, meaning the class with the highest probability of the sample belonging to that class's distribution. 
    
\myparagraph{Open-World Evaluation} In this case, unlike before, test samples from both seen and unseen classes are evaluated. At any incremental step $k$, all test samples belonging to the seen distribution, \ie $D_1, ..., D_k$ are considered ID, while any other classes outside the seen distribution: $D_{k+1}, .., D_T$, where $T$ is the last class, are considered \textit{far} OOD~\cite{yang2022openood}. We measure the discriminative power of the OOD detector by evaluating AUC and AUPR on scores received from IND and OOD samples.

Evaluation metrics are computed for each incremental model until the last task. Closed-World metrics are computed from the second to the last task since there will not be any OOD data after the last task.

\subsection{Results and Discussion}
\label{sec: results_disc}

Here, we discuss the closed-world and open-world performance for each method: \MoreFw, \MORE, \BUILD as meta-rows for datasets: \CifartenFiveT, \CifarhunTenT, \CifarhunTwenT
as meta-columns in \autoref{tab:full_results} and \MoreFw and \BUILD for \TimageNetFiveT and \TimageNetTenT in~\autoref{tab:tinyimg}. Each meta-row is further categorized into the scoring function (or scorer) considered in this work: SM, SMMD, EN, and ENMD, and then further into OOD detectors: Base, ReAct, DICE, and SCALE.


\begin{table*}[htbp] 
\centering 
\scriptsize
\resizebox{0.95\linewidth}{!}{ 
\begin{tabular}{lll|ccccc|ccccc}
\toprule

\multicolumn{3}{c|}{} 
& \multicolumn{5}{c|}{\TimageNetFiveT} 
& \multicolumn{5}{c}{\TimageNetTenT} \\

\cmidrule(lr){4-13}
\multicolumn{3}{c|}{} & \multicolumn{3}{c}{Closed-World} & \multicolumn{2}{c|}{Open-World} & \multicolumn{3}{c}{Closed-World} & \multicolumn{2}{c}{Open-World} \\

\cmidrule(lr){4-6} \cmidrule(lr){7-8} \cmidrule(lr){9-11} \cmidrule(lr){12-13}

 &  &  
& \textbf{\ACA}$\uparrow$ & \textbf{\AIA}$\uparrow$ & \textbf{\AF}$\downarrow$ & \textbf{\AUC}$\uparrow$ & \textbf{\AUPR}$\uparrow$ 
& \textbf{\ACA}$\uparrow$ & \textbf{\AIA}$\uparrow$ & \textbf{\AF}$\downarrow$ & \textbf{\AUC}$\uparrow$ & \textbf{\AUPR}$\uparrow$ \\

\midrule
\multirow{16}{*}{\rotatebox{90}{\textbf{\MoreFw}}} 

& \multirow{4}{*}{\rotatebox{90}{\textbf{SM}}}
 & \textbf{Base} & 59.87 & 69.48 & 5.17 & 76.81 & 88.92 & 56.48 & 65.31 & 4.01 & 77.16 & 93.21 \\
 & & \textbf{React} & 59.82 & 69.51 & 5.64 & 77.45 & 89.25 & 57.11 & 65.97 & 4.1 & 77.3 & 93.33 \\
 & & \textbf{Dice} & 54.34 & 65.38 & 4.7 & 75.45 & 87.82 & 50.29 & 61.24 & 3.38 & 75.6 & 92.56 \\
 & & \textbf{Scale} & 59.97 & 69.57 & 5.23 & 76.96 & 88.98 & 56.52 & 65.28 & 3.96 & 77.27 & 93.22 \\
\cline{3-13}
& \multirow{4}{*}{\rotatebox{90}{\textbf{SMMD}}}
 & \textbf{Base} & 63.43 & 71.51 & 6.04 & 81.21 & 91.29 & 61.9 & 69.92 & 4.97 & 81.31 & 94.88 \\
 & & \textbf{React} & \textbf{63.61} & \textbf{71.69} & \textbf{6.19} & 81.37 & 91.4 & 62.02 & 69.89 & 4.94 & 81.2 & 94.88 \\
 & & \textbf{Dice} & 59.08 & 68.41 & 4.9 & 80.68 & 91.24 & 56.16 & 66.09 & 3.82 & 81.12 & 94.94 \\
 & & \textbf{Scale} & 63.43 & 71.5 & 6.14 & 81.31 & 91.34 & \textbf{61.97} & \textbf{69.95} & \textbf{4.91} & 81.43 & 94.91 \\
\cline{3-13}
& \multirow{4}{*}{\rotatebox{90}{\textbf{EN}}}
 & \textbf{Base} & 44.2 & 58.5 & 29.72 & 84.53 & 93.31 & 37.3 & 49.95 & 30.36 & 78.89 & 94.11 \\
 & & \textbf{React} & 45.69 & 59.74 & 27.39 & 84.27 & 93.16 & 39.28 & 52.45 & 28.66 & 78.3 & 93.93 \\
 & & \textbf{Dice} & 28.31 & 48.15 & 42.95 & 79.49 & 90.65 & 17.47 & 34.88 & 42.24 & 73.3 & 92.26 \\
 & & \textbf{Scale} & 43.66 & 58.26 & 30.52 & 84.7 & 93.38 & 37.06 & 49.92 & 30.66 & 79.11 & 94.21 \\
\cline{3-13}
& \multirow{4}{*}{\rotatebox{90}{\textbf{ENMD}}}
 & \textbf{Base} & 60.6 & 69.3 & 13.56 & 86.65 & 94.33 & 59.97 & 67.41 & 12.33 & 84.17 & 96.18 \\
 & & \textbf{React} & 61.33 & 69.78 & 12.51 & 86.32 & 94.16 & 60.0 & 67.97 & 12.2 & 83.73 & 96.06 \\
 & & \textbf{Dice} & 48.5 & 61.44 & 25.72 & 84.21 & 93.15 & 47.48 & 58.81 & 22.59 & 81.09 & 95.32 \\
 & & \textbf{Scale} & 60.36 & 69.26 & 13.8 & \textbf{86.9} & \textbf{94.43} & 59.65 & 67.24 & 12.44 & \textbf{84.47} & \textbf{96.25} \\
\bottomrule
 \rowcolor{lightblue}
 & & \textbf{Avg.} & 54.76 & 65.72 & 15.01 & 81.77 & 91.68 & 51.29 & 61.39 & 14.1 & 79.72 & 94.39 \\
\bottomrule
\bottomrule

\multirow{16}{*}{\rotatebox{90}{\textbf{\BUILD}}}

& \multirow{4}{*}{\rotatebox{90}{\textbf{SM}}}
 & \textbf{Base} & 56.85 & 66.93 & 11.69 & 77.5 & 89.05 & 52.36 & 62.57 & 11.39 & 76.58 & 93.14 \\
 & & \textbf{React} & 56.92 & 67.04 & 11.55 & 77.85 & 89.28 & 52.75 & 63.06 & 11.37 & 76.99 & 93.35 \\
 & & \textbf{Dice} & 52.45 & 64.05 & 13.31 & 76.3 & 88.07 & 48.53 & 60.1 & 12.84 & 75.45 & 92.68 \\
 & & \textbf{Scale} & 56.84 & 66.98 & 11.84 & 77.64 & 89.11 & 52.43 & 62.63 & 11.42 & 76.73 & 93.18 \\
\cline{3-13}
& \multirow{4}{*}{\rotatebox{90}{\textbf{SMMD}}}
 & \textbf{Base} & 62.23 & 70.18 & 10.18 & 81.59 & 91.36 & 60.97 & 68.95 & 9.44 & 80.94 & 94.77 \\
 & & \textbf{React} & 62.04 & 70.36 & 10.28 & 81.69 & 91.45 & 60.56 & 68.59 & 9.47 & 81.01 & 94.82 \\
 & & \textbf{Dice} & 58.61 & 68.23 & 11.41 & 81.08 & 91.1 & 57.6 & 66.53 & 10.34 & 81.0 & 94.84 \\
 & & \textbf{Scale} & 62.14 & 70.2 & 10.3 & 81.65 & 91.39 & 60.9 & 68.89 & 9.53 & 81.04 & 94.79 \\
\cline{3-13}
& \multirow{4}{*}{\rotatebox{90}{\textbf{EN}}}
 & \textbf{Base} & 64.64 & 71.98 & 8.21 & 85.96 & 93.7 & 61.73 & 69.88 & 10.18 & 84.47 & 96.05 \\
 & & \textbf{React} & 64.41 & 71.75 & 8.01 & 85.64 & 93.55 & 61.59 & 69.74 & 9.99 & 83.91 & 95.9 \\
 & & \textbf{Dice} & 58.71 & 68.38 & 12.59 & 83.18 & 92.25 & 56.17 & 66.32 & 12.69 & 81.88 & 95.24 \\
 & & \textbf{Scale} & 64.66 & 71.96 & 8.27 & 86.0 & 93.68 & 61.82 & 70.02 & 10.28 & 84.6 & 96.06 \\
\cline{3-13}
& \multirow{4}{*}{\rotatebox{90}{\textbf{ENMD}}}
 & \textbf{Base} & \textbf{66.34} & \textbf{72.84} & \textbf{7.98} & 87.29 & 94.38 & \textbf{64.56} & \textbf{71.82} & \textbf{9.19} & 85.76 & 96.53 \\
 & & \textbf{React} & 66.2 & 72.74 & 7.65 & 86.99 & 94.24 & 64.28 & 71.76 & 9.14 & 85.27 & 96.39 \\
 & & \textbf{Dice} & 62.76 & 70.98 & 11.24 & 86.19 & 93.81 & 61.07 & 69.43 & 10.67 & 84.73 & 96.21 \\
 & & \textbf{Scale} & 66.29 & 72.8 & 8.01 & \textbf{87.42} & \textbf{94.43} & 64.6 & 71.79 & 9.09 & \textbf{85.95} & \textbf{96.58} \\
\bottomrule
 \rowcolor{lightblue}
 & & \textbf{Avg.} & 61.38 & 69.84 & 10.16 & 82.75 & 91.93 & 58.87 & 67.63 & 10.44 & 81.64 & 95.03 \\
\bottomrule

\end{tabular}
}
\caption{Results for model: \MoreFw, and BUILD on the Tiny-Imagenet dataset, using different scorers and detectors, along with closed-world and open-world metrics. 
For each model and dataset, the row with the highest AIA (ACA in case of ties) is highlighted in bold for ACA, AIA, and AF, while the row with the highest AUC (AUPR in case of ties) is highlighted in bold for AUC and AUPR.
}
\label{tab:tinyimg} 
\end{table*}

\subsubsection{\MORE: Baseline Performance}
The reproduced results from the original work correspond to the SMMD scores for the Base detector in the tables, which obtained better performances on the smallest dataset \Cifarten and lower performances on the larger datasets. 
In a closed-world setting, considering \ACA and \AIA without \AF and vice versa in isolation can be misleading. \ACA and \AIA suggest how well the model performs after learning new classes, \AF sheds light on how much the model forgets by the end since it first learns the task. For example, considering EN scores for all detectors \ACA and \AIA is higher for \CifartenFiveT compared to \CifarhunTenT and \CifarhunTwenT. Still, at the same time, \AF is much higher for \CifartenFiveT, suggesting the model's inability to perform correct task-id prediction on that dataset. Even if the model suffers in closed-world settings using EN scores, the detectors can still detect unseen samples, giving high \AUC and \AUPR performance. \MORE performs best using SMMD scores.

\subsubsection{\MORE vs \MoreFw}

After evaluating these models in closed- and open-world settings, we discovered that \MoreFw seems to perform better than the version proposed by the authors of that work (\MORE) in many configuration settings, suggesting that additional training may, in fact, be unnecessary. We observed that, in SM and SMMD cases, \MoreFw gives a comparable or superior performance to \MORE and does not suffer as much when using EN or ENMD scores. In fact, \AF in this setting is the best across all methods. When using Base or SCALE as the OOD detector and SMMD or ENMD as the scoring function, we can achieve good performance in this setting.

\subsubsection{\BUILD Performance}
On average, \BUILD gives the best results and is stable across all the scoring functions. The framework achieves its highest effectiveness with the EN and ENMD scoring functions, yielding the lowest \AF\ in these configurations. Nonetheless, \BUILD\ does not consistently outperform \MORE\ and its variant in the closed-world setting, particularly when employing the SM and SMMD scoring functions. When \AF is the biggest concern, the best model is \MoreFw; however, it takes longer to train, and it leverages the buffer, which, as we explained in the previous sections, creates privacy issues and may facilitate different adversarial attacks. \BUILD performs consistently well on all detectors and scorers when compared to \MORE and \MoreFw in the open-world setting and also in the closed-world setting when considering \ACA and \AIA. These findings are further corroborated by results on additional datasets, as presented in~\autoref{tab:tinyimg}, where we highlighted in bold closed-world performance with the best performing AIA and open-world performance with the best AUC.

\subsubsection{Rejection Performance}
\begin{figure*}[!htbp]
    \centering
    \includegraphics[width=\linewidth]{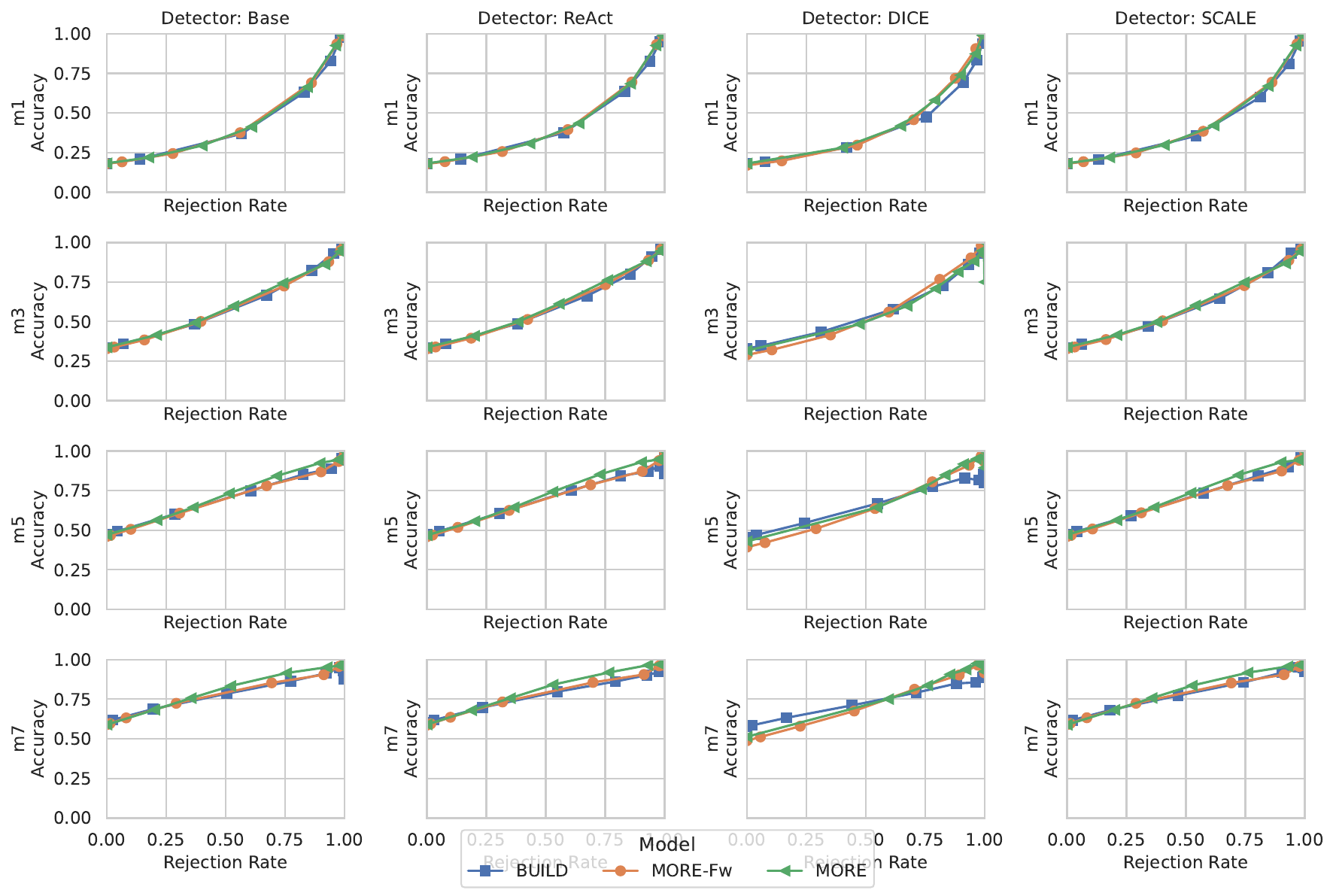}
    
    \caption{Accuracy vs Rejection Curve for Cifar100-10T: Threshold-free evaluation of OOD detectors corresponding to best scores for each model, \ie SMMD for \MORE and \MoreFw, and ENMD for \BUILD. For each plot, the x-axis is the rejection rate, and the y-axis is the accuracy of the model trained after the 1st, 3rd, 5th, and 7th task row-wise.} 
    \label{fig:rejection_cifar100_enmd_smmd}
\end{figure*}

The central objective of detecting OOD samples is to ensure that unseen samples are not misclassified into known classes, enabling their rejection.\autoref{fig:rejection_cifar100_enmd_smmd} illustrates this through accuracy–rejection curves on \CifarhunTenT across varying thresholds, with rejection rate on the x-axis and classification accuracy on the y-axis. Each row represents an incrementally trained model, and each column corresponds to a different OOD detector. Within each plot, we compare \MoreFw, \MORE, and \BUILD using their best-performing scoring functions (SMMD for \MORE and \MoreFw, ENMD for \BUILD). For example, in $m_1$ the model has seen 20 classes and evaluated on all 100, resulting in an IND:OOD ratio of 1:4. Therefore, we see an initial accuracy of about 20\% at 0\% rejection. As the rejection threshold increases, more samples are filtered out, leading to higher accuracy since only retained samples are classified. Similar trends are observed for subsequent models: $m_3$, $m_5$, $m_7$. Overall, \BUILD achieves performance comparable to its buffered counterparts, \MORE and \MoreFw.

\subsubsection{Post-hoc Performance}

Across datasets, detector performance remains closely matched for all evaluation metrics. Base and SCALE emerge as the most effective detectors, while DICE underperforms, likely due to its masking of weight vectors, unlike ReAct and SCALE, which apply activation clipping or scaling. In contrast, scoring functions exhibit more pronounced differences. Softmax-based scores (SM, SMMD) are optimal for \MoreFw\ and \MORE, whereas energy-based scores (EN, ENMD) yield the best results for \BUILD. Augmenting SM and EN with Mahalanobis Distance (MD) generally improves multi-head model performance in both closed- and open-world settings, though gains are limited for single-head models (explained later in \autoref{sec: single-head method}). While MD requires centroids from training activations and incurs additional computational cost, it remains advantageous for smaller datasets. Energy-based scoring underperforms for \MORE methods despite its common use in post-hoc detection. This can be attributed to the exposure of \MORE to OOD data during training, which biases activation clipping/scaling, and therefore, reduces reliability. In contrast, \BUILD\ avoids this bias by training exclusively on IND data, explaining its superior ENMD performance. Softmax, while suboptimal for OOD detection, benefits from buffer optimization in \MORE, enabling improved task separation in its default setting. Although the MD calculated for both \MORE and \BUILD models is affected by the presence or absence of the buffer, its role as a supporting function to SM and EN consistently enhances performance in multi-head architectures.

\begin{figure*}
    \centering
    \includegraphics[width=0.49\linewidth]{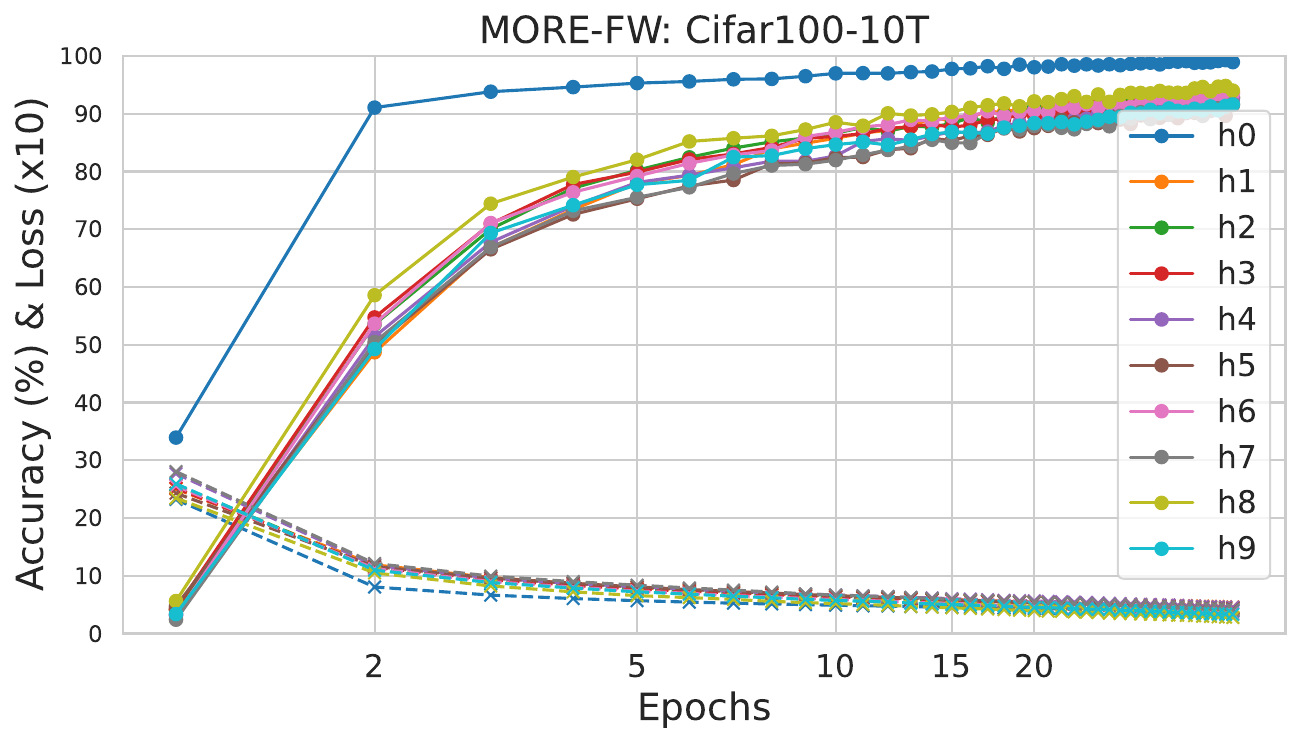}
    \includegraphics[width=0.49\linewidth]{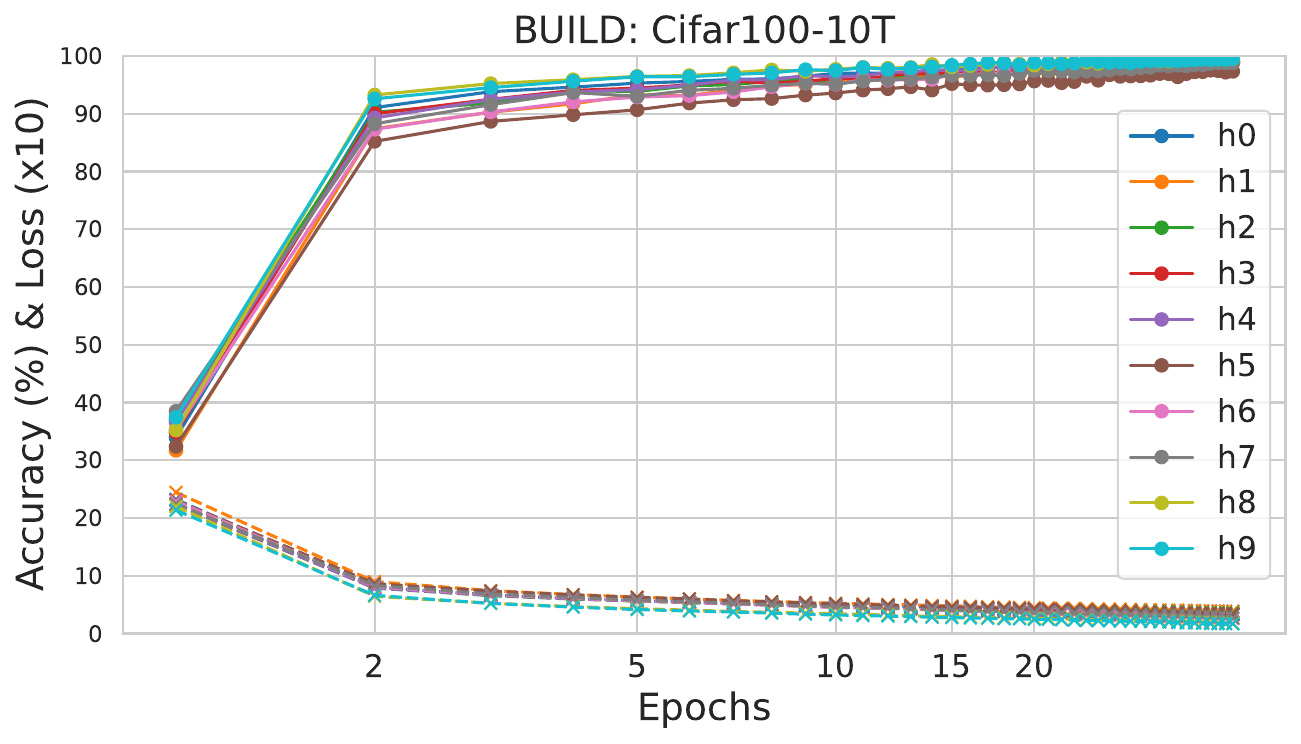}
    \caption{Training accuracy and loss for \CifarhunTenT for \MORE-Fw and \BUILD. Note here, the left plot is after the forward pass, to get the final training, backupdates does additional training.}
    \label{fig: train_perf}
\end{figure*}

\subsubsection{Train Time Compute}
\begin{figure*}[t]
    \centering
    \includegraphics[width=\linewidth]{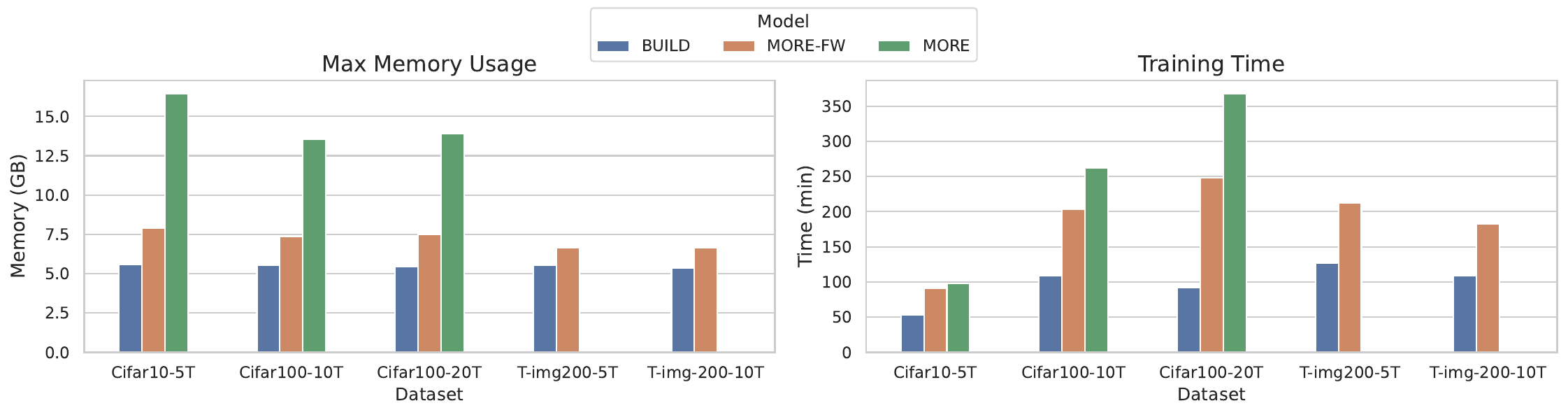}
    \caption{(a) Max Memory Usage and (b) Total training time, evaluated for each model and dataset}
    \label{fig: memory_train_comp}
\end{figure*}

Memory buffers contribute a lot to the large memory consumption of the models. For example, raw images from CIFAR datasets of size $32\times32\times3$ in a buffer of size 2000 are approximately 6.1M vector values, similarly for \Timgnet, it's 24.6M $(64\times64\times3)$ vector values. Processing 6.1M or 24.6M elements in each incremental training step for \MoreFw and additionally again during backupdate step in \MORE leads to almost double the Maximum Memory Usage (MMU) for \MoreFw and more than triple for \MORE compared to \BUILD has no buffer (see \autoref{fig: memory_train_comp}). We also observed that buffered methods take longer to achieve the desired training accuracy than their buffer-free counterparts. For example, when training \CifarhunTenT in ~\autoref{fig: train_perf}, \BUILD reaches the 90\% training accuracy at the fifth epoch, whereas it takes around 40 epochs for \MoreFw to achieve similar performance.
While \MORE and \MoreFw give good performance in certain settings, they both are trained using a buffer and, consequently, utilize more compute resources and take longer to train, as shown in \autoref{fig: memory_train_comp}. \BUILD takes around one-third (half) of the memory usage and half the training time for \Cifarten (\Cifarhun) compared to \MORE, and a similar memory usage but less than half of the training time compared to \MoreFw.


\subsubsection{Test Time Compute}
\begin{figure*}[htbp]
    \centering
    \includegraphics[width=0.5\linewidth]{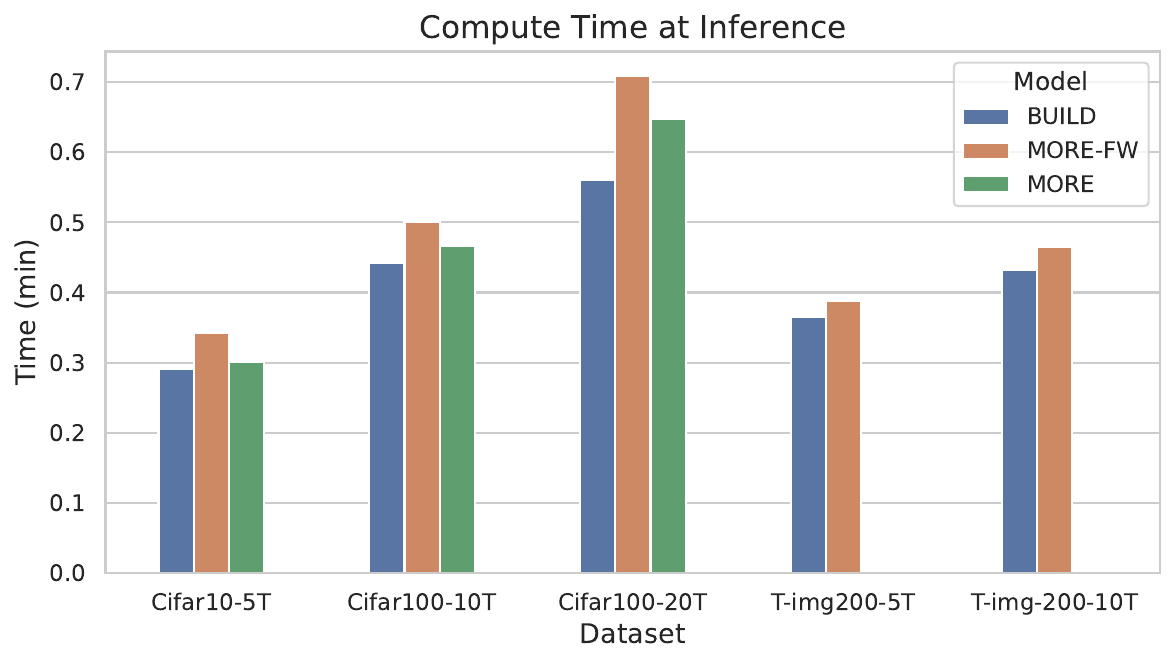}
    \caption{Total compute time of OOD scores calculated for ReAct evaluated for each multi-head model}
    \label{fig: inf-time compute}
\end{figure*}

At inference time, all models compute OOD scores for each head to choose the desired head. The time complexity grows as the number of heads grows, which is an inherent challenge in multi-head models.~\autoref{fig: inf-time compute} shows the time taken to compute OOD scores at the last trained model, for example, for \Cifarten, the computed time is after training $m4$ and consequently $m9$ for \CifarhunTenT and $m19$ for \CifarhunTwenT. For the sake of analysis, we computed the ReAct energy scores. For comparative analysis, we see \BUILD takes relatively less time to predict the class.  We observe that the least difference between \BUILD and \MORE methods is about 2 seconds for 2000 samples, which scales rapidly for large chunks of test data. In terms of resource compute, both \MORE and \BUILD take the same memory at inference time.

\subsection{Ablation on Adapter Size}
Adapters are key to enabling efficient task learning. Following prior work, we use an adapter size of 64, but also conduct an ablation study on \Cifarten using \BUILD with adapter sizes {16, 32, 64, 128}.~\autoref{tab: adapter_size} presents results with SCALE as the OOD detector and ENMD as the scorer. Reducing the size from 64 to 16 yields an ACA decrease of 0.44 (~0.48\%) and an AUC drop of ~0.47\%, while increasing to 128 gives an ACA gain of ~0.65\% and an AUC gain of ~0.25\%. Since overall performance remains largely stable, a smaller adapter size offers a more efficient choice.

\begin{table*}[ht]

\centering
\scriptsize
\begin{tabular}{c|ccccc}
\toprule
\textbf{Adapter} & \textbf{ACA}$\uparrow$ & \textbf{AIA}$\uparrow$ & \textbf{AF}$\downarrow$ & \textbf{AUC}$\uparrow$ & \textbf{AUPR}$\uparrow$ \\
\textbf{Size} & & & & & \\
\midrule
16   & 89.42 (\textit{-0.44}) & 93.49 (\textit{-0.21}) & 3.02 (\textit{+0.63}) & 89.97 (\textit{-0.43}) & 95.55 (\textit{+0.02}) \\
32   & 89.10 (\textit{-0.76}) & 93.19 (\textit{-0.51}) & 3.18 (\textit{+0.47}) & 89.76 (\textit{-0.64}) & 95.46 (\textit{-0.07}) \\
64   & 89.86             & 93.70             & 3.65              & 90.40             & 95.53 \\
128  & 90.45 (\textit{+0.59}) & 94.02 (\textit{+0.32}) & 3.15 (\textit{+0.5}) & 90.63 (\textit{+0.23}) & 95.80 (\textit{+0.27}) \\
\bottomrule
\end{tabular}
\caption{Performance across adapter sizes shown as x($\Delta$), with $\Delta$ denoting change from baseline 64. +/- indicates improvement/decline. Results are for \BUILD on \Cifarten with ENMD+SCALE. }
\label{tab: adapter_size}
\end{table*}

\begin{table*}[!htbp] 
\centering 
\resizebox{0.99\linewidth}{!}{
\begin{tabular}{lll|ccccc|ccccc|ccccc}
\toprule

\multicolumn{3}{c|}{} 
& \multicolumn{5}{c|}{\CifartenFiveT}
& \multicolumn{5}{c|}{\CifarhunTenT}
& \multicolumn{5}{c}{\CifarhunTwenT} \\
\cmidrule(lr){4-18}

\multicolumn{3}{c|}{} &
\multicolumn{3}{c}{Closed-World} & \multicolumn{2}{c|}{Open-World} &
\multicolumn{3}{c}{Closed-World} & \multicolumn{2}{c|}{Open-World} &
\multicolumn{3}{c}{Closed-World} & \multicolumn{2}{c}{Open-World} \\

\cmidrule(lr){4-6} \cmidrule(lr){7-8} \cmidrule(lr){9-11} \cmidrule(lr){12-13} \cmidrule(lr){14-16} \cmidrule(lr){17-18}

 &  &  
& \textbf{\ACA}$\uparrow$ & \textbf{\AIA}$\uparrow$ & \textbf{\AF}$\downarrow$ & \textbf{\AUC}$\uparrow$ & \textbf{\AUPR}$\uparrow$ 
& \textbf{\ACA}$\uparrow$ & \textbf{\AIA}$\uparrow$ & \textbf{\AF}$\downarrow$ & \textbf{\AUC}$\uparrow$ & \textbf{\AUPR}$\uparrow$ 
& \textbf{\ACA}$\uparrow$ & \textbf{\AIA}$\uparrow$ & \textbf{\AF}$\downarrow$ & \textbf{\AUC}$\uparrow$ & \textbf{\AUPR}$\uparrow$  \\

\midrule
\multirow{16}{*}{\rotatebox{90}{\textbf{DER++}}} 
& \multirow{4}{*}{\rotatebox{90}{\textbf{SM}}}
 & \textbf{Base} & 84.12 & 89.26 & 18.65 & 86.21 & 92.57 & \textbf{66.47} & \textbf{80.3} & \textbf{28.9} & 78.32 & 93.29 & \textbf{69.27} & \textbf{82.16} & \textbf{25.27} & 78.6 & 96.35 \\
 & & \textbf{React} & 83.25 & 89.23 & 18.58 & \textbf{89.07} & \textbf{93.58} & 64.99 & 78.55 & 28.41 & 77.56 & 92.73 & 68.24 & 80.97 & 25.27 & 78.54 & 96.22 \\
 & & \textbf{Dice} & 16.15 & 40.72 & 5.35 & 51.41 & 79.33 & 10.98 & 26.68 & 12.33 & 50.55 & 84.65 & 5.19 & 16.4 & 6.03 & 40.81 & 87.78 \\
 & & \textbf{Scale} & 83.37 & 89.12 & 18.51 & 88.76 & 93.13 & 64.69 & 78.56 & 28.53 & 77.8 & 92.8 & 68.25 & 80.98 & 25.28 & 78.59 & 96.19 \\
\cline{3-18}
& \multirow{4}{*}{\rotatebox{90}{\textbf{SMMD}}}
 & \textbf{Base} & 53.83 & 57.41 & 56.61 & 60.58 & 84.81 & 28.79 & 47.6 & 74.14 & 46.94 & 84.38 & 29.71 & 51.44 & 71.8 & 37.19 & 87.69 \\
 & & \textbf{React} & 19.92 & 45.28 & 99.0 & 72.88 & 88.38 & 9.73 & 28.15 & 95.36 & 56.89 & 86.97 & 4.93 & 17.62 & 97.94 & 51.74 & 90.5 \\
 & & \textbf{Dice} & 19.91 & 45.27 & 98.96 & 53.19 & 81.77 & 9.73 & 28.12 & 95.28 & 51.7 & 85.4 & 4.93 & 17.61 & 97.89 & 41.9 & 87.91 \\
 & & \textbf{Scale} & 19.92 & 45.28 & 99.0 & 72.85 & 88.37 & 9.73 & 28.15 & 95.36 & 57.0 & 86.99 & 4.93 & 17.62 & 97.94 & 51.61 & 90.49 \\
\cline{3-18}
& \multirow{4}{*}{\rotatebox{90}{\textbf{EN}}}
 & \textbf{Base} & 19.82 & 45.35 & 0.14 & 86.63 & 92.05 & 9.52 & 28.37 & 0.22 & \textbf{80.41} & \textbf{93.7} & 4.68 & 17.44 & 0.21 & \textbf{80.96} & \textbf{96.49} \\
 & & \textbf{React} & \textbf{84.04} & \textbf{89.28} & \textbf{18.66} & 87.1 & 92.56 & 65.96 & 79.89 & 29.32 & 80.06 & 93.72 & 68.96 & 82.0 & 25.39 & 80.91 & 96.62 \\
 & & \textbf{Dice} & 61.67 & 78.65 & 41.58 & 81.56 & 89.84 & 18.5 & 38.98 & 35.32 & 75.04 & 91.49 & 17.75 & 35.27 & 27.42 & 77.48 & 95.49 \\
 & & \textbf{Scale} & 84.01 & 89.2 & 18.76 & 85.87 & 91.59 & 66.1 & 80.0 & 29.08 & 80.35 & 93.68 & 69.18 & 82.09 & 25.27 & 80.92 & 96.47 \\
\cline{3-18}
& \multirow{4}{*}{\rotatebox{90}{\textbf{ENMD}}}
 & \textbf{Base} & 19.91 & 45.28 & 99.01 & 84.53 & 92.78 & 9.72 & 28.13 & 95.37 & 75.9 & 92.33 & 4.93 & 17.62 & 97.94 & 77.4 & 95.92 \\
 & & \textbf{React} & 40.92 & 58.38 & 72.75 & 69.47 & 88.07 & 9.76 & 28.16 & 95.32 & 56.07 & 86.87 & 4.94 & 17.63 & 97.93 & 49.74 & 90.11 \\
 & & \textbf{Dice} & 19.91 & 45.28 & 99.01 & 79.49 & 91.29 & 9.72 & 28.13 & 95.37 & 67.11 & 89.69 & 4.93 & 17.62 & 97.94 & 70.32 & 94.63 \\
 & & \textbf{Scale} & 44.72 & 60.03 & 68.0 & 70.01 & 88.27 & 9.78 & 28.16 & 95.3 & 56.1 & 86.92 & 4.94 & 17.66 & 97.93 & 49.53 & 90.1 \\
\bottomrule
 \rowcolor{lightblue}
 & & \textbf{Avg.} & 47.22 & 63.31 & 52.04 & 76.23 & 89.27 & 29.01 & 46.0 & 58.35 & 66.74 & 89.73 & 27.24 & 40.76 & 57.34 & 64.14 & 93.06 \\
\bottomrule
\bottomrule

\multirow{16}{*}{\rotatebox{90}{\textbf{PASS}}} 
& \multirow{4}{*}{\rotatebox{90}{\textbf{SM}}}
 & \textbf{Base} & \textbf{83.07} & \textbf{89.22} & \textbf{12.55} & \textbf{82.54} & \textbf{92.32} & \textbf{70.65} & \textbf{79.46} & \textbf{14.7} & \textbf{79.82} & \textbf{93.56} & \textbf{69.17} & \textbf{78.66} & \textbf{16.92} & \textbf{78.71} & \textbf{95.96} \\
 & & \textbf{React} & 17.06 & 50.53 & 31.76 & 47.94 & 74.7 & 7.85 & 26.73 & 12.38 & 56.97 & 84.58 & 3.68 & 17.45 & 6.51 & 56.19 & 90.28 \\
 & & \textbf{Dice} & 21.85 & 45.9 & 1.09 & 47.26 & 75.1 & 8.98 & 28.39 & 15.18 & 49.92 & 81.18 & 4.99 & 18.08 & 9.43 & 46.81 & 87.5 \\
 & & \textbf{Scale} & 16.6 & 50.54 & 31.58 & 47.62 & 74.31 & 7.86 & 26.71 & 12.73 & 57.25 & 84.63 & 3.58 & 17.41 & 6.68 & 56.5 & 90.35 \\
\cline{3-18}
& \multirow{4}{*}{\rotatebox{90}{\textbf{SMMD}}}
 & \textbf{Base} & 24.41 & 46.43 & 91.8 & 66.48 & 86.46 & 14.92 & 33.22 & 87.48 & 56.54 & 86.73 & 11.46 & 26.31 & 89.22 & 53.1 & 90.98 \\
 & & \textbf{React} & 20.03 & 44.91 & 97.24 & 66.85 & 82.63 & 9.96 & 27.89 & 91.08 & 56.59 & 83.97 & 5.18 & 17.74 & 93.6 & 54.63 & 89.64 \\
 & & \textbf{Dice} & 19.94 & 44.95 & 97.21 & 51.02 & 77.69 & 9.88 & 27.82 & 92.24 & 53.53 & 82.87 & 5.13 & 17.72 & 94.56 & 49.74 & 88.47 \\
 & & \textbf{Scale} & 20.06 & 44.92 & 97.2 & 66.71 & 82.62 & 9.94 & 27.87 & 90.88 & 56.62 & 83.97 & 5.12 & 17.74 & 93.53 & 54.54 & 89.6 \\
\cline{3-18}
& \multirow{4}{*}{\rotatebox{90}{\textbf{EN}}}
 & \textbf{Base} & 19.84 & 45.31 & 0.09 & 67.4 & 86.61 & 9.35 & 27.63 & 0.18 & 67.26 & 89.46 & 4.74 & 17.22 & 0.06 & 68.13 & 93.84 \\
 & & \textbf{React} & 15.73 & 53.67 & 39.38 & 47.12 & 76.14 & 11.74 & 29.27 & 30.44 & 55.33 & 84.03 & 4.56 & 18.91 & 16.38 & 56.66 & 90.56 \\
 & & \textbf{Dice} & 52.54 & 73.76 & 26.22 & 67.35 & 84.59 & 22.84 & 45.05 & 31.76 & 56.72 & 83.73 & 18.76 & 38.24 & 25.62 & 58.07 & 90.64 \\
 & & \textbf{Scale} & 16.15 & 54.1 & 41.85 & 46.5 & 75.59 & 11.5 & 29.26 & 29.39 & 55.61 & 84.17 & 4.5 & 18.98 & 16.15 & 57.12 & 90.7 \\
\cline{3-18}
& \multirow{4}{*}{\rotatebox{90}{\textbf{ENMD}}}
 & \textbf{Base} & 19.97 & 44.9 & 97.35 & 77.39 & 88.88 & 9.98 & 27.84 & 93.0 & 64.48 & 88.36 & 5.24 & 17.78 & 95.77 & 62.11 & 92.45 \\
 & & \textbf{React} & 19.67 & 50.45 & 67.04 & 49.93 & 76.76 & 9.96 & 27.81 & 92.81 & 57.77 & 84.92 & 5.13 & 17.8 & 95.08 & 53.92 & 89.76 \\
 & & \textbf{Dice} & 20.0 & 44.9 & 97.31 & 71.84 & 86.07 & 9.99 & 27.85 & 92.99 & 59.8 & 86.52 & 5.24 & 17.78 & 95.77 & 56.59 & 90.93 \\
 & & \textbf{Scale} & 19.72 & 50.45 & 65.12 & 50.09 & 76.75 & 9.95 & 27.8 & 92.78 & 57.63 & 84.78 & 5.12 & 17.81 & 94.8 & 53.76 & 89.69 \\
\bottomrule
\rowcolor{lightblue}
 & & \textbf{Avg.} & 25.42 & 52.18 & 55.92 & 59.63 & 81.08 & 14.71 & 32.54 & 55.0 & 58.86 & 85.47 & 10.1 & 23.48 & 53.13 & 57.29 & 90.71 \\
\bottomrule

\end{tabular}
}
\caption{Results DER++ and PASS on the CIFAR-10 and CIFAR-100 datasets, using different scorers and detectors, along with closed-world and open-world metrics. 
For each model and dataset, the row with the highest AIA (LCA in case of ties) is highlighted in bold for LCA, AIA, and AF, while the row with the highest AUC (AUPR in case of ties) is highlighted in bold for AUC and AUPR.
}
\label{tab:single_heads} 
\end{table*}

\subsection{Applying BUILD to Single-Head Methods}
\label{sec: single-head method}
We now extend our evaluation to include two competitive non-multi-head CIL methods, namely DER++~\cite{buzzega2020derpp} and PASS~\cite{Zhu2021pass}, and assess their classification accuracy while recognizing never-before-seen classes.
DER++ is a low-resource replay-based method that distills knowledge from stored samples of past tasks to improve retention while learning new tasks. We followed the experimental setup used for MORE for the buffer size, and set the method-specific hyperparameters $\alpha=0.3$ and $\beta=0.5$.
On the other hand, PASS is buffer-free, storing a single class prototype for each past class and applying prototype augmentation in the deep feature space, combined with self-supervised learning to enhance robustness. PASS required longer training to converge; we set the number of epochs to 50 for CIFAR-10 and 100 for CIFAR-100. The learning rate was fixed to $10^{-3}$ for both datasets. Following the original recommendations and further preliminary tuning, we set the knowledge distillation weight to $0.1$, the prototype augmentation weight to $10$, and the temperature during training to $0.1$. To stabilize optimization, we also applied gradient clipping to an $L_2$ norm of $10$.
All other experimental conditions were kept consistent with our main experimental protocol to ensure a fair comparison.

In \autoref{tab:single_heads}, we find that both DER++ and PASS suffer performance degradation—under closed- and open-world metrics—when paired with detection methods outside their original inference strategy, namely plain maximum softmax without feature adjustment (SM + Base). Even in this setting, their performance remains below the best results of \BUILD. For Mahalanobis-based scores (SMMD and ENMD), such drops are expected in single-head models: after each incremental step, the model has no incentive to preserve class-specific feature representations, causing class centroids to drift and reducing the reliability of distance-based measures. EN scores also perform poorly in this context, as they are not strong indicators of IND classification~\cite{sun2021react}, especially in single-head models where no task-ID predictions are required. Nevertheless, DER++ can partially recover when supported with buffered class data and refined by post-hoc feature adjustments, whereas with no buffer data, PASS collapses entirely outside its base setting. Overall, these results underscore the robustness and adaptability of \BUILD across diverse scoring strategies.

\section{Related Work}
\label{sec: related_work}

Although there is a wealth of work in CL and OOD domains individually, there are few works that have adopted TIL methods to perform CIL with task-id prediction in an open-world setting. \MORE \cite{kim2022more} employs OOD detection for task identification, replaying samples during forward training before updating task heads. ROW \cite{kim2023learnability} follows a similar design, adding theoretical guarantees of CIL learnability, while TPL \cite{lin2024TPL} adopts \MORE’s training scheme but replaces the OOD score with a likelihood ratio. CEDL \cite{aguilar2023continual} incorporates deep evidential learning \cite{sensoy2018_DeepEvidential} into buffer-based continual learning. Despite operating in open-world settings, all these methods rely on replay buffers and do not evaluate rejection performance—a primary objective of OOD detection. 

Only a few incremental learning (IL) approaches employ low- or no-memory strategies. PNN~\cite{rusu2022pnn} and DER~\cite{yan2021derdynamic} expand a new backbone for each task while freezing previous ones, leading to substantial memory demands—particularly with transformer architectures. In contrast, we adopt parameter-efficient adapters \cite{houlsby2019parameterefficient} to learn new classes while mitigating forgetting. Although PASS~\cite{Zhu2021pass} avoids storing exemplars, it retains class prototypes and operates in a single-head setting, where its best performance remains below the average performance of BUILD (\autoref{tab:single_heads}). Furthermore, none of these methods address OOD detection at deployment. Few OOD detection methods integrate IL in their pipeline. OOD-UCL \cite{he2022_ood_ucl} corrects output bias in unsupervised CL, OSIL \cite{leo2019_osil} clusters novel classes for continual training on text data, and KNNENS \cite{zhang_2023_knnens} detects novel classes via $k$-nearest neighbors before storing them in a buffer. Most treat IL as a post-detection step with stored samples in memory, diverging from IL’s assumption of no access to past data and the need to prevent forgetting. In this work, we detect OOD samples while learning new classes. \BUILD performs OOD detection and evaluates the accuracy of the model based on the rejection ratio of the detected samples, giving a realistic view of model deployment in the wild.

\section{Conclusion, Limitations and Future Work}\label{sec: conclusions}

Continual learning in real-world scenarios must operate under open-world conditions. This work evaluates \MORE, a state-of-the-art open-world CIL method, against \BUILD, a novel buffer-free approach that eliminates weight updates. Experiments show that \BUILD trains substantially faster while maintaining comparable performance, and it outperforms other low-resource baselines such as DER++ and PASS. Moreover, \BUILD offers greater stability across scoring functions and avoids the privacy and security risks associated with replay buffers.

Although \BUILD enables efficient training and alleviates privacy concerns by removing raw data buffers, it encounters scalability challenges with a large number of tasks due to the computational cost of multi-head architectures. While multi-head models help reduce catastrophic forgetting, some forgetting still occurs, largely from the growing output space rather than from limitations of the approach itself. Future work will include a broader evaluation of OOD detectors, the incorporation of far-OOD and adversarial datasets, and enhancements to inference-time efficiency. We also aim to explore the applicability of \BUILD to other modalities such as text and audio.

\section*{Acknowledgments}
This work has been partly supported by the EU-funded Horizon Europe projects ELSA (GA no. 101070617), Sec4AI4Sec (GA no. 101120393), and CoEvolution (GA no. 101168560); and by the projects SERICS (PE00000014) and FAIR (PE00000013) under the MUR National Recovery and Resilience Plan funded by
the European Union - NextGenerationEU.
This work was conducted while Srishti Gupta was enrolled in the Italian National Doctorate on AI run by Sapienza University of Rome in collaboration with the University of Cagliari.

\biboptions{sort&compress}
\bibliographystyle{elsarticle-num} 

\end{document}